\documentclass[journal]{IEEEtran}
\usepackage{cite}
\usepackage{url}
\usepackage{hyperref}
\hypersetup{colorlinks,linkcolor={black},citecolor={black},urlcolor={black}}  
\usepackage{bm}
\usepackage{amssymb}
\usepackage{amsmath}
\usepackage{amsthm}

\usepackage{graphicx}
\usepackage[caption=false,font=normalsize,labelfont=sf,textfont=sf]{subfig}
\usepackage{booktabs}
\usepackage{multirow}
\usepackage{bbding}
\usepackage{algorithm}
\usepackage{algpseudocode}
\usepackage{adjustbox}
\usepackage{balance}
\usepackage[marginal]{footmisc}
\usepackage{color}
\usepackage{hyperref}
\hypersetup{
    colorlinks=true,
    urlcolor=blue,
}
\begin{document}
\title{Picking Up Quantization Steps for Compressed Image Classification}

\author{Li~Ma,
        Peixi~Peng,
        Guangyao~Chen,
        Yifan~Zhao,
        Siwei~Dong,
        Yonghong~Tian,~\IEEEmembership{Fellow, IEEE}
        \thanks{Corresponding author: Peixi Peng and Yonghong Tian}
        \thanks{Li Ma, Peixi Peng, Guangyao Chen, Yifan Zhao and Yonghong Tian are with the School of Computer Science, Peking University, Beijing 100871, China, (e-mail: mali\_hp@pku.edu.cn; pxpeng@pku.edu.cn; gy.chen@pku.edu.cn; zhaoyf@pku.edu.cn; yhtian@pku.edu.cn).}
        \thanks{Peixi Peng, Yonghong Tian are also with Network Intelligence Research, PengCheng Laboratory, Shenzhen 518066, China}
        \thanks{Yonghong Tian is also with the School of Electronics Computer Engineering, Peking University, Shenzhen 518055, China.}
        \thanks{Siwei Dong is with Beijing Academy of Artificial Intelligence, Beijing 100083, China, (email: swdong@baai.ac.cn).}
}

\markboth{}%
{}


\maketitle

\begin{abstract}
The sensitivity of deep neural networks to compressed images hinders their usage in many real applications, which means classification networks may fail just after taking a screenshot and saving it as a compressed file. In this paper, we argue that neglected disposable coding parameters stored in compressed files could be picked up to reduce the sensitivity of deep neural networks to compressed images. Specifically, we resort to using one of the representative parameters, quantization steps, to facilitate image classification. Firstly, based on quantization steps, we propose a novel quantization aware confidence (QAC), which is utilized as sample weights to reduce the influence of quantization on network training. Secondly, we utilize quantization steps to alleviate the variance of feature distributions, where a quantization aware batch normalization (QABN) is proposed to replace batch normalization of classification networks. Extensive experiments show that the proposed method significantly improves the performance of classification networks on CIFAR-10, CIFAR-100, and ImageNet. The code is released on \href{https://github.com/LiMaPKU/QSAM.git}{https://github.com/LiMaPKU/QSAM.git}
\end{abstract}


\section{Introduction}
\label{sec:intro}

\IEEEPARstart{{D}}{{EEP}}
neural networks have shown superior performance on a number of machine vision tasks, especially in Image classification \cite{shu2015weakly,tang2016generalized,He2016Deepresnet,tan2019efficientnet,regnetradosavovic2020designing,yang2020rotation,tan2021efficientnetv2}. However, recent papers have reported that the performance of image classification significantly drops under the existence of lossy compression \cite{liu2020comprehensive,endo2020classifying,pei2019effects,endo2021cnnbased,ma2021reducing,wan2020featureconsistency,zhang2021just,wei2022effects,ghosh2018robustness}, \emph{i.e.}, deep neural networks are sensitive to lossy compressed images. As shown in Fig. \ref{fig:imagenet_samples}, the network makes drastically different classifications of the images compressed with JPEG \cite{wallace1992jpeg}. That is, deep neural networks may fail just after you take a screenshot and save it. Therefore, it is essential to eliminate the sensibility caused by compression loss for deploying deep neural networks on various applications.
\begin{figure}[t]
\centering
\includegraphics[width=0.90\linewidth]{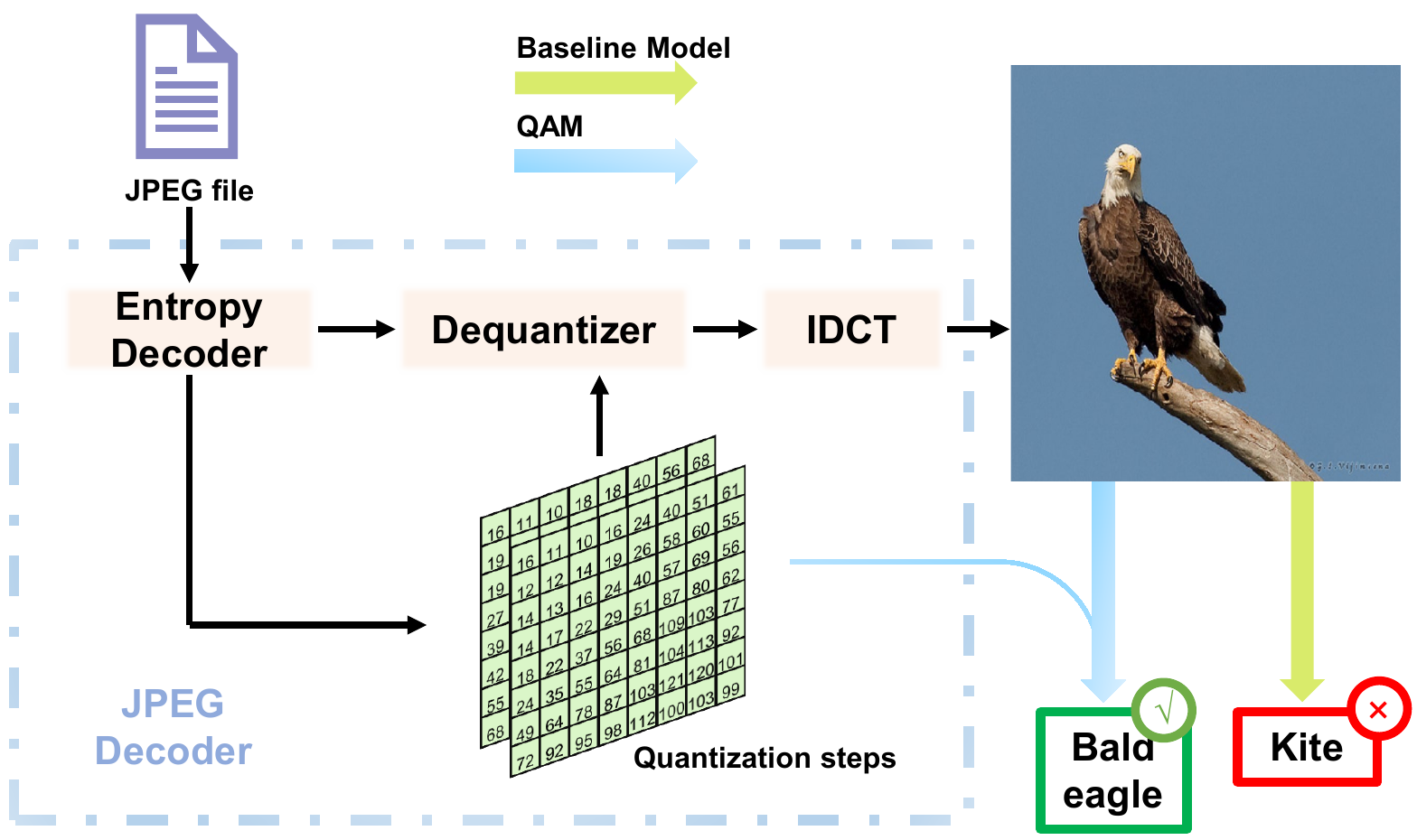}
\caption{The illustration of the proposed quantization aware method. The baseline model trained by only pixel values of images is sensitive, leading to some behaviors counter-intuitive to humans. The proposed quantization aware method, taking quantization steps into account, has a better generalization ability for compressed images.}
\label{fig:ill_qam}
\vspace{-3mm}
\end{figure}

Numerous methods have been proposed to tackle this problem, which roughly fall into three categories: data augmentation-based methods, ensemble learning-based methods, and feature alignment-based methods. Data augmentation methods treat compression as one kind of data augmentation and re-train or fine-tune classification networks on compressed images. However, these methods ignore the correlation between original images and compressed images and could not handle the compressed images whose compression levels vary in a wide range \cite{dodge2017quality,zhou2017classification}. Another type of methods utilizes ensemble learning, which trains several classification networks and design strategies to adjust the ensemble weights \cite{endo2020classifying,endo2021cnnbased}. Since ensemble learning-based methods require much more computation than single classification networks, we focus on single network approaches in this work. Recently, some methods \cite{wan2020featureconsistency,ma2021reducing} reduce the sensitivity of deep neural networks to compressed images by aligning features produced by original and compressed images. These methods focus on heavily compressed images, whose JPEG quality factors are below 40, and are not concerned about slightly compressed images. However, as shown in Figure \ref{fig:imagenet_samples}, slight compression could mislead the network to make a different classification. Moreover, these feature alignment-based methods need original images with annotations as inputs. However, plenty of datasets are already compressed, including ImageNet \cite{deng2009imagenet}, PASCAL VOC \cite{everingham2015pascal}, Market-1501 \cite{market1501}, and MSMT17 \cite{msmt17}, \emph{etc.} To these issues, we are interested in this work if there are efficient ways to reduce the sensitivity of deep neural networks to compressed images without the need for original images. Considering quantization is the main reason causing compression loss, we focus on quantization steps and model lossy compression in image classification to reduce the sensitivity of deep neural networks, as shown in Fig. \ref{fig:ill_qam}. The quantization steps are one of the key parameters in lossy compression, and amount of studies \cite{hung1991optimal,wu1993rate,crouse1997joint,ratnakar2000efficient,yang2014quantization,wang2001designing,jiang2011jpeg,hopkins2017simulated,chao2013design,tuba2014jpeg,tuba2017jpeg,duan2012optimizing,yan2020qnet,li2021quantization,yang2021aclustering,lin2011apassive,ik1999adaptive} explore how to design or estimate the quantization steps.
\begin{figure*}[t]
\centering
\includegraphics[width=0.85\linewidth]{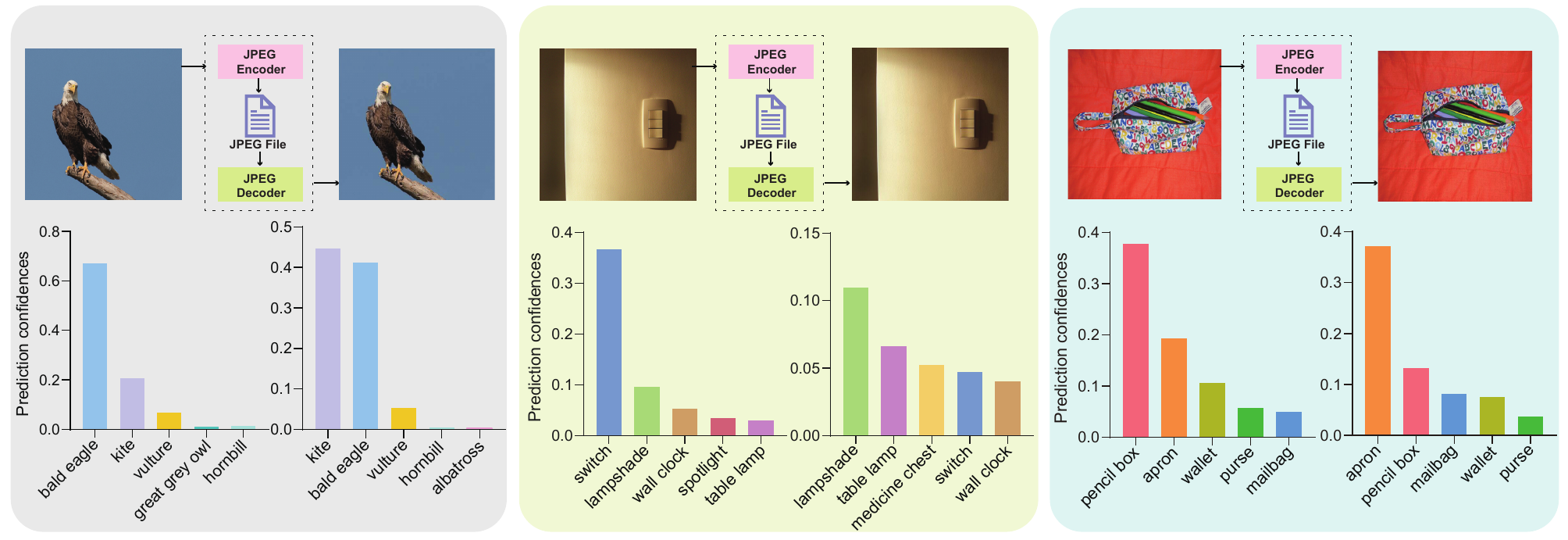}
\vspace{-2mm}
\caption{Three testing samples selected from ImageNet that show the counter-intuitive behaviors to compressed images: the model (ResNet50) correctly predicts the original images but makes drastically different classifications of the images compressed slightly by JPEG. The prediction confidences are also shown. The JPEG encoder and decoder employed in this experiment are the standard ones in the Python Image Library (PIL) with the default quality factor, 75.}
\label{fig:imagenet_samples}
\vspace{-3mm}
\end{figure*}
Moreover, the quantization steps have been proven to be effective on several visual tasks. For example, Kornblum \emph{et al.} \cite{kornblum2008using} utilized quantization steps to identify whether computers have altered images. Park \emph{et al.} \cite{park2018double} design a deep neural network for double JPEG detection using quantization steps. Besides, quantization steps are utilized to assist the artifacts reduction \cite{zhao2017reducing,liu2019graphbased,feng2019coding,ouyang2020towards,ehrlich2020quantization,mu2020graphbased}. However, the relationship between quantization steps and image classification remains unexplored. As a many-to-one mapping, quantization could change images in detail, probabilistically interfering with network training. Moreover, various kinds of quantization steps could cause the shift of feature distributions, which does not meet the intrinsic assumption of batch normalization and leads to performance degradation. To address these two issues, we propose a quantization steps aware method (QAM), composed of two modules, \emph{i.e.}, the quantization steps aware confidence module, and the quantization steps aware batch normalization module. The former aims to reduce the influence of detail change on network training, while the latter aims to alleviate the influence of feature distribution misalignment. Firstly, since quantization is irreversible, it could change images in texture and structure. Compressed images whose important details are changed by quantization could interfere with network training. To address this problem, we conduct a theoretical analysis in the frequency domain and propose quantization steps aware confidence (QAC) to measure the influence of quantization on network training. During network training, we weight training samples with QAC to reduce the influence of quantization. Secondly, since quantization steps determine how many bits are used to encode the components on each frequency band, images with different kinds of quantization steps indicate distinct distributions of their features, which do not meet the intrinsic assumption of batch normalization. We propose a quantization-aware batch normalization (QABN) to address this problem, which utilizes distribution bases and a meta-learner to predict the target distribution.

To verify the effectiveness of QAM, we conduct various experiments based on CIFAR-10 \cite{cifar}, CIFAR-100 \cite{cifar} and ImageNet \cite{deng2009imagenet}. Considering a typical classification model, QAM achieves a significant improvement in classification performance by weighting training samples with QAC and replacing batch normalization with QABN. Moreover, QAM shows a strong generalization ability to other compression formats. Our major contributions are summarized as follows:
\begin{itemize}
    \item To the best of our knowledge, we are the first to utilize quantization steps to reduce the sensitivity of deep neural networks to compressed images and improve the performance of image classification.
    \item We model the relationship between quantization steps and network training, and propose quantization steps aware confidence as sample weights to help networks alleviate the influence of quantization.
    \item To alleviate the influence of feature distribution misalignment, we propose a quantization aware batch normalization, which utilizes quantization steps and distribution bases to predict the target distribution.
    \item Extensive experiments demonstrate the effectiveness of the proposed method. With the help of quantization steps, we could improve 6.3\% and 1.2\% accuracy on CIFAR-100-J and ImageNet-J, respectively.
\end{itemize}

The rest of this paper is organized as follows. Section \ref{sec:rw} reviews related works and Section \ref{sec:pm} describes the proposed quantization aware method. Experiments and analysis of results are presented in Section \ref{sec:exp}. Section \ref{sec:con} finally concludes this paper.

\section{Related Work}
\label{sec:rw}
\subsection{Classification of Compressed Images}
Image classification is a fundamental and important problem in computer vision \cite{shu2016image,lu2007survey,haralick1973textural,boiman2008defense,He2016Deepresnet,tan2019efficientnet,regnetradosavovic2020designing,tan2021efficientnetv2,tang2016generalized,shu2015weakly,yang2020rotation}. Although deep neural networks have achieved superior performance, their sensitivity to compressed images has been discovered recently \cite{liu2020comprehensive,endo2020classifying,pei2019effects,endo2021cnnbased,ma2021reducing,wan2020featureconsistency,zhang2021just,wei2022effects,ghosh2018robustness}. To eliminate the sensibility to compressed images, numerous methods have been proposed, primarily data augmentation-based methods, ensemble learning-based methods, and feature alignment-based methods. Data augmentation-based methods regard compression as one type of image degradation and treat the degraded images as new training samples \cite{dodge2017quality,zhou2017classification}. However, these methods ignore the correlation between original images and their compressed version and could not handle the images with various compressed levels. Another type of approaches utilizes ensemble learning. Ghosh \emph{et al.} \cite{ghosh2018robustness} propose an ensemble network and a method based on maximum a posteriori to make the final decision. Endo \emph{et al.}  \cite{endo2020classifying} exploit classification networks trained on restored and original images, and calculate ensemble weights according to the estimated compressed level to improve the classification performance. Then, they utilize the features of the classification network trained on restored images as cues to determine ensemble weights \cite{endo2021cnnbased}. Ensemble learning-based methods require much more computation than a single classification network, which is a bottleneck. Feature-based methods eliminate the sensibility by aligning the features extracted from original and compressed images. Wan \emph{et al.} \cite{wan2020featureconsistency} design a parallel pipeline to train the classification network, where they utilize a feature consistency constraint to guide the training of the network. Besides the feature-level constraint, Ma \emph{et al.} \cite{ma2021reducing} propose a pixel-level constraint to reinforce the feature alignment ability. Although these methods achieve good performance, they need original images with annotations as inputs. Since plenty of datasets are already compressed, we focus on reducing the sensitivity of deep neural networks to compressed images without the need for original images.

\subsection{Quantization Steps}
\label{sec:jpeg}
Quantization is a general process to save bits in lossy compression. Generally, quantization is applied to transformation coefficients based on the pre-defined quantization steps that determine how many bits are used to encode the components on each frequency band. Since the human visual system is less able to distinguish high-frequency components, the quantization steps on high-frequency are generally larger than the ones on low-frequency components. Generally, an $h \times w \times c$ image has $hwc$ quantization steps. For simplicity, we form these quantization steps into an $h\times w\times c$ tensor, named quantization steps tensor (QST) and denoted as $Q$ in this paper.

\begin{figure}[t!]
\centering
\includegraphics[width=0.90\linewidth]{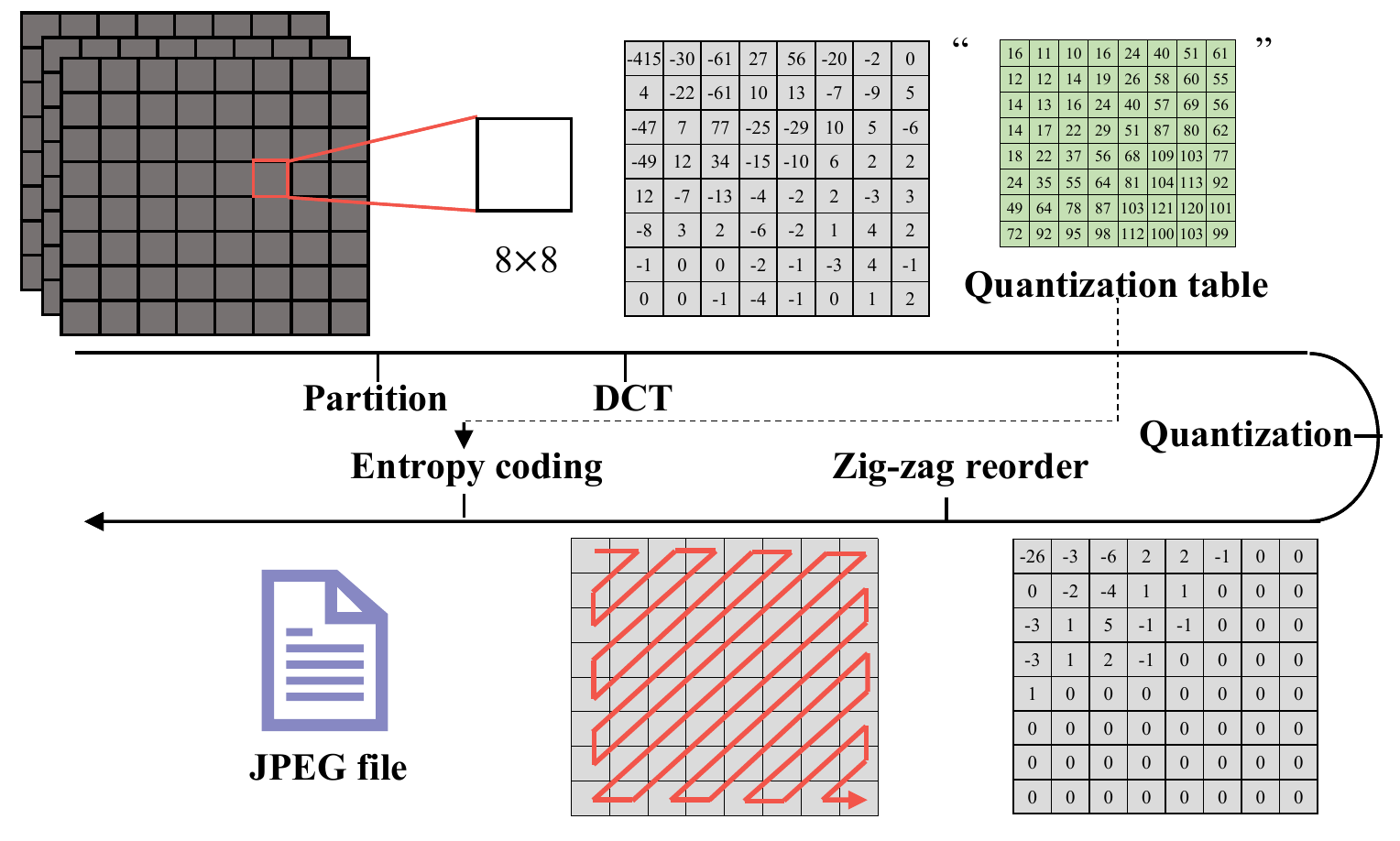}
\caption{A brief overview of the JPEG compression technology. The quantization steps are specified by a 64-element quantization table.}
\label{fig:fig2}
\vspace{-3mm}
\end{figure}
JPEG\cite{wallace1992jpeg} is one of the most widespread still images compression standards and composes a base part of generally used video compression \cite{wiegand2003overview,sullivan2012overview,gao2014overview,zhang2019recent,chen2018overview,mukherjee2013latest,bross2021vvc,jiang2021adistortion,lin2021modulated}. Fig. \ref{fig:fig2} shows a brief overview of JPEG compression technology. Firstly, each spatial component is partitioned into $8\times 8$ non-overlapping blocks. Secondly, transformations, \emph{e.g.} discrete cosine transformation (DCT), follow to generate 64 transformation coefficients. Thirdly, the transformation coefficients are quantized based on quantization steps: $c^{'}_{i,j}=round[\frac{c_{i,j}}{q_{i,j}}]$, here $c_{i,j}$ is the transformation coefficient and 64-element $[q_{i,j}]_{8\times 8}$ are the pre-defined quantization table. Fourthly, the quantized coefficients are serialized using a zig-zag order. Finally, the reordered coefficients and the quantization table are losslessly compressed with entropy coding. JPEG standard provides example quantization tables (called the default quantization tables in this paper), and users could use the default quantization tables and adjust the quality factor (QF, ranging from 0 to 100) to obtain 101 kinds of quantization tables. Amount of studies show that the quantization table plays an important role in JPEG compression performance. Therefore, many methods are proposed to design quantization tables, such as rate-distortion optimization \cite{hung1991optimal,wu1993rate,crouse1997joint,ratnakar2000efficient}, human visual system-based optimization \cite{wang2001designing,jiang2011jpeg}, heuristic optimization \cite{hopkins2017simulated}, and vision tasks-based optimization \cite{chao2013design,tuba2014jpeg,tuba2017jpeg,duan2012optimizing,yan2020qnet,liu2018deepn}. Generally, the quantization table for luminance components is different from that for chrominance components, and each $8 \times 8$ block shares two 64-element quantization tables. For simplicity, we concatenate them together to get an $8\times8\times 2$ tensor and utilize the tensor as the QST of JPEG. Besides JPEG, some coding formats may not store quantization steps directly, and we could still calculate the quantization steps by other coding parameters.

\subsection{Batch Normalization}
Ioffe \emph{et al.} \cite{ioffe2015batch} introduce Batch normalization (BN) to accelerate networks' training by permitting to use higher learning rates and reducing the sensitivity to network initialization. Since introduced, BN has been proven to be a critical element in the successful training of ever-deeper neural architectures \cite{yang2019mean, bjorck2018understanding, luo2018towards, garbin2020dropout}. When utilizing BN, one inherent assumption is that the input features should come from a single or similar distribution. Recently, some researchers have modified the traditional BN to cover problems where the input has different underlying distributions \cite{xie2020adversarial,tsai2020robust}. To counter the mixed distributions, Xie \emph{et al.} \cite{xie2020adversarial} propose a normalization layer that keeps separate BNs to features belonging to clean and adversarial samples. Tsai \emph{et al.} \cite{tsai2020robust} propose a noise-aware calibration in batch normalization statistics, which effectively rectifies the shifted distribution caused by the noise during analog computing. These methods concern different distributions but pay no attention to their relationship. On the other hand, there are many QSTs, making it impractical to keep separate BNs for each QST. Unlike these methods, the proposed QABN models the relationship between feature distributions and quantization steps, and predicts the target BN depending on the corresponding quantization steps.

\section{Proposed Method}
\label{sec:pm}
\subsection{Problem Formulation}
Let $\mathfrak{C}(x,Q)$ denote the compressed image whose QST is $Q=(q_{i,j,c})_{8\times 8 \times 2}$. $\mathcal{X}$, $\mathcal{Y}$ and $Q$ denote the image space, label space, and QST, respectively. $(\mathfrak{C}(x,Q),y) \in \mathcal{X}\times \mathcal{Y}$ denotes a sample-label pair from the trainset $\mathcal{D}$. $f(\cdot;\theta_{f})$ denotes a baseline model (\emph{e.g.} ResNet-50\cite{He2016Deepresnet}) whose parameters are denoted as $\theta_{f}$. Ignoring lossy compression, we optimize the baseline model on training data by minimizing a cross-entropy loss:
\begin{equation}
\label{eq:baseline}
	{\min}\sum_{(\mathfrak{C}(x,Q),y) \in \mathcal{D}} {\mathcal{L}_{ce}(f(\mathfrak{C}(x,Q);\theta), y)}).
\end{equation}
Here, $\mathcal{L}_{ce}(\cdot, \cdot)$ is the cross-entropy loss function. However, trained by only pixel values of compressed images, the baseline model could be misled by quantization and suffer from the misalignment of feature distributions. To address these issues, we analyze how quantization affects the network training by exploring the correlations of images with different compression levels in the training. On this basis, we propose quantization aware confidence to guide the training and quantization aware batch normalization to align the distributions. 
\subsection{Quantization Aware Confidence}
As mentioned in Section \ref{sec:intro}, lossy compression could change images in texture and structure. Trained on these compressed images, the classification network may be misled. To quantitatively analyze the influence of quantization on network training, we analyze the training process in the frequency domain. Without losing generality, we consider the luminance channel of a compressed image $\mathfrak{C}(x,Q)$. Assuming $p_{m,n}$ as a single pixel, $p_{m,n}$ could be represented by $8 \times 8$ discrete cosine transformation in JPEG decompression:
\begin{equation}
\label{idct}
p_{m,n}=\sum_{i=0}^{7} \sum_{j=0}^{7} c^{'}{(m, n, i, j)} \cdot b{(i, j)},
\end{equation}
where $c^{'}{(m, n, i, j)}$ is the quantized transformation coefficient and $b(i,j)$ is the corresponding basis function at 64 different frequencies. When training a network $f(\cdot;\theta)$, the loss function to train the baseline model is calculated:
\begin{equation}
\label{eq:loss}
  \mathcal{L}_{base} = {\mathcal{L}_{ce}(f(\mathfrak{C}(x,Q);\theta), y)}).
\end{equation}
However, $\mathcal{L}_{base}$ is calculated from $\mathfrak{C}(x,Q)$ that contains compression loss. To measure how reliable $\mathcal{L}_{base}$ is, we consider the contribution of the component on $b(i,j)$. The gradient of $\mathcal{L}_{base}$ with respect to $b(i,j)$ could be calculated as:
\begin{align}
\label{BP_bij}
\frac{\partial \mathcal{L}_{base}}{\partial b{(i, j)}}&=\frac{\partial \mathcal{L}_{base}}{\partial f}\times\frac{\partial f}{\partial b{(i, j)}} \notag\\
&=\frac{\partial \mathcal{L}_{base}}{\partial f}\times \frac{\partial f}{\partial p_{m,n}} \times \frac{\partial p_{m,n}}{\partial b{(i, j)}} \notag\\
&=\frac{\partial \mathcal{L}_{base}}{\partial f}\times \frac{\partial f}{\partial p_{m,n}} \times c^{'}{(m, n, i, j)}.
\end{align}
Equation (\ref{BP_bij}) means that the contribution of the frequency component on $b{(i, j)}$ of the single pixel $p_{m,n}$ to train a network could be primarily determined by three component: the derivative of $\mathcal{L}_{base}$ with respect to $f$, the associated transformation coefficient ${c^{'}{(m, n, i, j)}}$, and the importance of the pixel $\frac{\partial f}{\partial p_{m,n}}$. Here $\frac{\partial \mathcal{L}_{base}}{\partial f}$ is fixed, and $\frac{\partial f(x; \theta)}{\partial p_{m,n}}$ is obtained after back-propagation\cite{hecht1992theory}, while ${c^{'}{(m, n, i, j)}}$ is distorted by quantization before training:
\begin{equation}
\label{eq:quantization}
    c^{'}{(m, n, i, j)} = round[\frac{c{(m, n, i, j)}}{q_{i,j,l}}] \times {q_{i,j,l}}.
\end{equation}
When $q_{i,j,l} \neq 1$, the frequency component on $b(i,j)$, which may contain details important to network training, may not be learned precisely.
Therefore, we take the approximate probability, $P(c^{'}{(m, n, i, j)} = c{(m, n, i, j)}) \approx \frac{1}{q_{i,j,l}}$ as the confidence of $b(i,j)$ on a single pixel $p_{m,n}$. Then, we take the average confidence of pixels as the quantization aware confidence of an image:
\begin{align}
\label{eq:qac}
    \text{QAC}(\mathfrak{C}(x,Q)) &= \frac{1}{mn} \sum_{m,n} \sum_{i,j,c} \frac{1}{q_{i,j,c}} \notag\\
    &= \sum_{i,j,c}\frac{1}{q_{i,j,c}}.
\end{align}
Finally, we utilize QAC as sample weights to calculate the quantization aware confidence-based cross-entropy loss:
\begin{equation}
 \mathcal{L}_{qac} = \sum_{(\mathfrak{C}(x,Q),y) \in \mathcal{D}} \text{QAC}(\mathfrak{C}(x,Q))\cdot {\mathcal{L}_{ce}(f(\mathfrak{C}(x,Q);\theta), y)}.
\end{equation}
Compared with normal cross-entropy loss, the quantization aware confidence-based cross-entropy loss utilizes different sample weights to compressed images with different compression levels. In other words, QAC utilizes quantization steps to model the relationship among images with different compression levels. On this basis, the quantization aware confidence-based cross-entropy loss assigns the contributions of images with different QSTs according to QAC, which could help the network trained on these compressed images alleviate the influence of quantization.

\begin{figure}[t!]
\centering
\includegraphics[width=.95\linewidth]{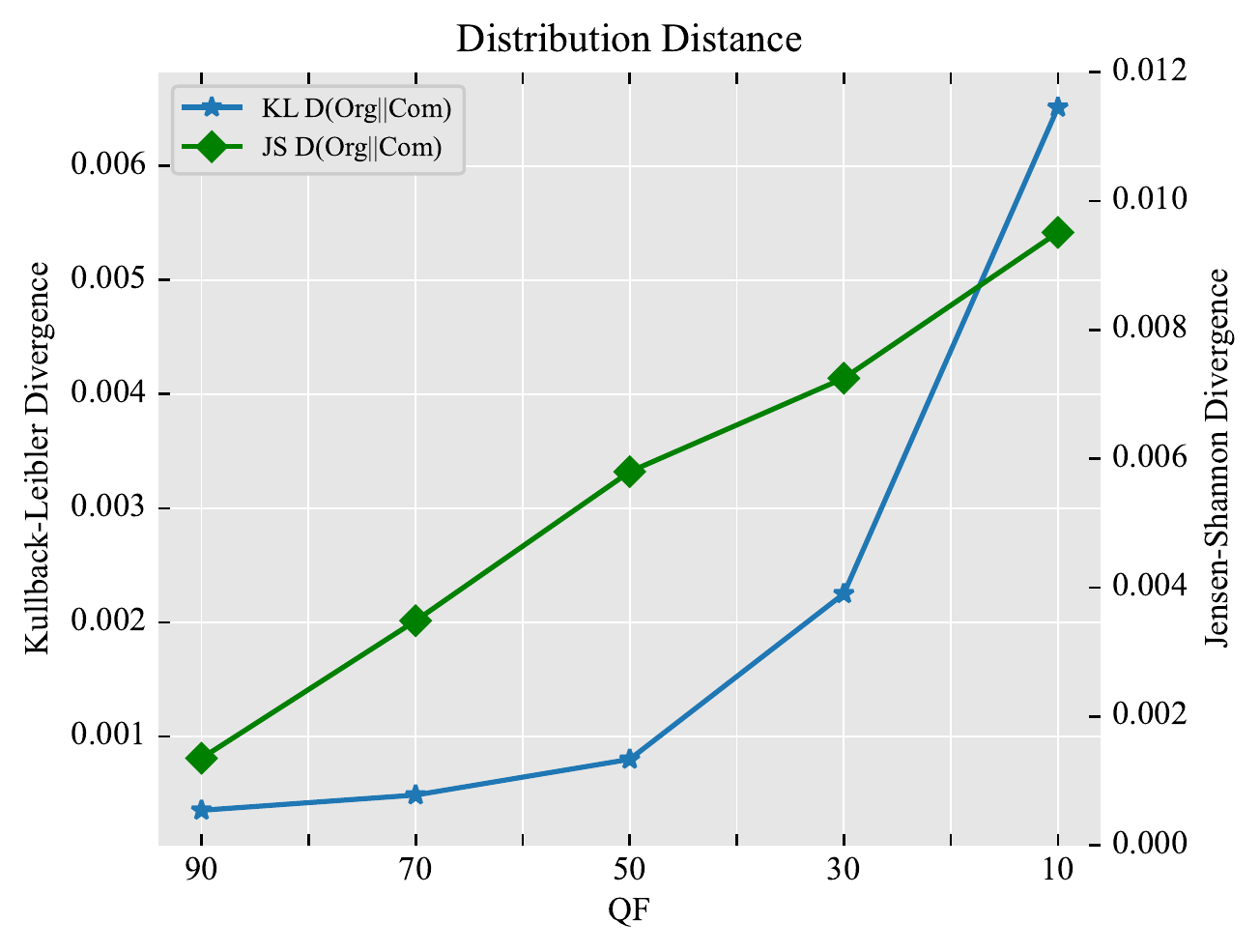}
\caption{Distribution distance between features learned from images with various QSTs. Here, we utilize the default quantization tables and adjust the quality factor to obtain various QSTs. The features represent the inputs of the first BN in ResNet-18. In this analysis, we consider two widely used distance functions, namely, Kullback-Leibler and Jensen-Shannon divergence, and they are calculated by Information Theoretical Estimators in Python \cite{itebox}.}
\label{fig:distance}
\vspace{-2mm}
\end{figure}
\begin{figure*}[t]
\centering
\includegraphics[width=0.90\linewidth]{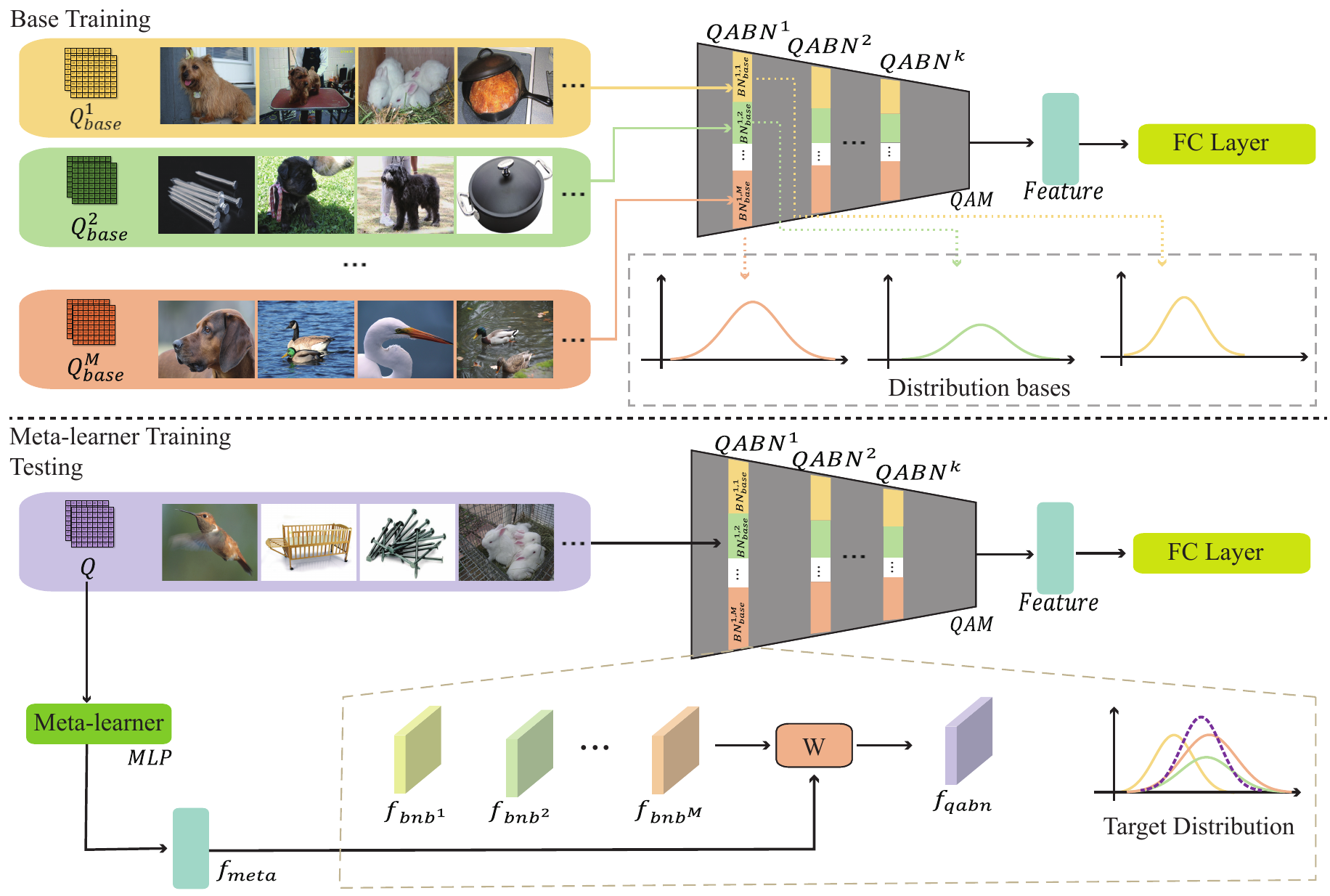}
\caption{The schematic diagram of quantization aware batch normalization. $QABN^{k}$ denotes $k$-th normalization layer in QAM. Each QABN contains $M$ parallel BNs, which are denoted as $BN_{base}^{k,1},BN_{base}^{k,2},...,BN_{base}^{k,M}$. In the base training process, features belonging to $Q^{i}_{base}$ are only normalized by $BN^{k,i}_{base}$. In the meta-learner training and testing process, features are normalized by all BNs in each QABN, which produce a group of normalized features: $f_{bnb^{1}}, f_{bnb^{2}}, ..., f_{bnb^{M}}$. Meanwhile, the meta-learner learns a quantization steps dependent feature ($f_{meta}$) whose size is $1 \times M$ from the QST $Q$. Finally, QABN takes the weighted sum of $f_{bnb^{1}}, f_{bnb^{2}}, ..., f_{bnb^{M}}$ with $F_{meta}$ as the weight. It should be noted that all QABNs in QAM share the same meta-learner.}
\label{fig:QABN}
\vspace{-2mm}
\end{figure*}

\subsection{Quantization Aware Batch Normalization}
One intrinsic assumption of BN is that the input features should come from a single or similar distribution \cite{xie2020adversarial, tsai2020robust}. However, as Fig. \ref{fig:distance} shows, features produced by images with various QSTs do not meet the assumption. 
\begin{figure}[t]
\centering
\includegraphics[width=0.8\linewidth]{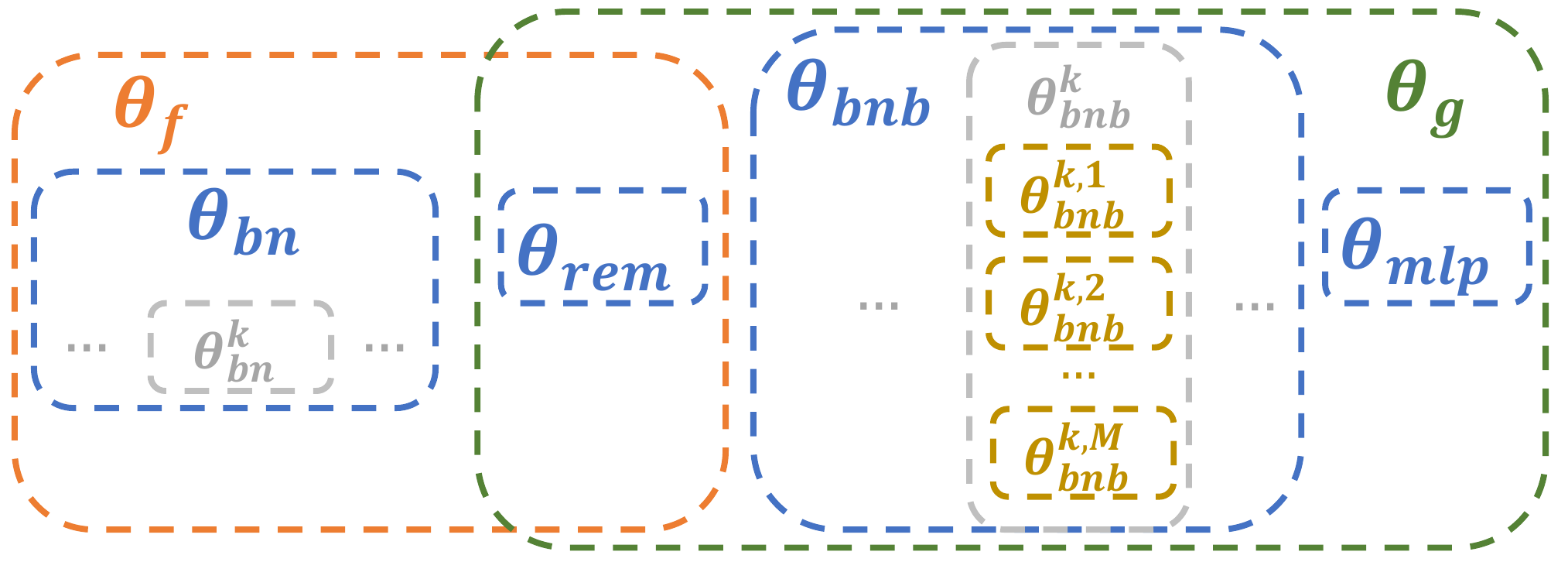}
\caption{The relationship between the parameters of the baseline model ($f(\cdot;\theta_f)$) and the proposed method ($g(\cdot;\theta_g)$).}
\label{fig:theta}
\vspace{-3mm}
\end{figure}

Similar to recent researches \cite{xie2020adversarial, tsai2020robust, xuan2021intra}, a straightforward method is keeping separate BNs to features that have different distributions. However, each element (\emph{i.e.} each quantization step) in QST can be any number from $[1, 255]$ \cite{wallace1992jpeg}, which means that there are $255^{128}$ QSTs theoretically. Therefore, we are unaware of which QSTs the networks will meet in real applications, making it impractical to keep separate BNs to different QSTs. To handle this problem, we propose QABN (shown in Fig. \ref{fig:QABN}), which utilizes quantization steps and distribution bases to predict the target distribution. Each QABN contains $M$ parallel BNs (called BN bases in this paper) whose parameters are denoted as $\theta^{k}_{bnb}= \{\theta^{k,1}_{base},\theta^{k,2}_{base},...,\theta^{k,M}_{base}\}$ and a multi-layer perceptron-based meta-learner $m(\cdot; \theta_{meta})$. In this manner, the distribution corresponding to a certain QST would be represented by BN bases.

The parameters of the baseline model $f(\cdot; \theta_f)$ can be divided into 2 parts: $\theta_f = \{\theta_{bn}, \theta_{rem}\}$. $\theta_{bn}$ denotes the parameters of all BNs in the baseline model, while $\theta_{rem}$ denotes the other parameters. The QAM could be denoted as $g(\cdot; \theta_{g})$, where the parameters of QAM could be divided into 3 parts: $\theta_{g} = \{\theta_{bnb}, \theta_{rem}, \theta_{meta}\}$. All QABNs in QAM share one meta-learner $m(\cdot; \theta_{meta})$. Fig. \ref{fig:theta} shows the relationship between parameters.

\subsection{Training Strategy}
QAM is trained by a two-step method consisting of base training and meta-learner training.

In the base training, we train $\theta_{bnb}$ and $\theta_{rem}$. Let $(\mathfrak{C}(x,Q),y) \in \mathcal{X}\times \mathcal{Y}$ denote a image-label pair from the trainset $\mathcal{D}$. Here, $\mathcal{X}$, $\mathcal{Y}$ denote the image space and label space, respectively; $Q$ denotes the QST of the compressed image. Firstly, we statistic the QSTs in the trainset. Secondly, depending on the QSTs distribution, we select some QSTs as QST bases. Let $\{Q^{1}_{base},Q^{2}_{base}, ... ,Q^{M}_{base} \}$ denote the selected $M$ QST bases. Thirdly, we divide $\mathcal{D}$ into $\mathcal{D}_{base}$ and $\mathcal{D}_{meta}$. Specifically, considering a image-label pair $(\mathfrak{C}(x,Q),y)$ from $\mathcal{D}$, if $Q$ belongs to $\{Q^{1}_{base},Q^{2}_{base}, ... ,Q^{M}_{base} \}$, we place $(\mathfrak{C}(x,Q),y)$ in $\mathcal{D}_{base}$, otherwise in $\mathcal{D}_{meta}$. Fourthly, we optimize $\theta_{bnb}$ and $\theta_{rem}$ on $\mathcal{D}_{base}$ by minimizing the quantization aware confidence-based cross-entropy loss:
\begin{equation}
\label{eq:base_train1}
    \{\hat{\theta}_{bnb}, \hat{\theta}_{rem}\} =
    \underset {\{\theta_{bnb}, \theta_{rem}\}}{\text{arg min}}
    \mathcal{L}_{qac}(x,y,g(\cdot;\{\theta_{bnb},\theta_{rem}\})).
\end{equation}
In this process, features belonging to $Q^{i}_{base}$ are only normalized by $i$-th BN in each QABN. Therefore, the parameters of $i$-th BN in each QABN are only optimized on images whose QST is $Q^{i}_{base}$. By contrast, the parameters of BNs in the baseline model are optimized on all training samples.

In the meta-learner training, we train $\theta_{meta}$ on $\mathcal{D}_{base}$ and $\mathcal{D}_{meta}$. As shown in Fig. \ref{fig:QABN}, the meta-learner learns a quantization steps dependent feature $f_{meta}$ whose size is $1\times M$. Based on $f_{meta}$, we perform a feature-wise weighted sum on the output of BN bases to generate the normalized features $f_{qabn}$. It should be noted that we are unaware of which QSTs the networks will meet in real applications. In other words, in real applications, networks may meet QSTs that are unseen in training. To improve the generalization ability of QAM to unseen QSTs, inspired by meta-learning, we design a gradient-based meta-learning scheme, where we simulate meeting unseen QSTs in the training process to promote optimization. The foundation of the gradient-based meta-learning scheme is simulating meeting new QSTs in the training process to promote performance when meeting new QSTs in the testing. Specifically, each iteration could be divided into two sub-steps. Firstly, we sample a mini-batch $X_{base},Y_{base}$ from $\mathcal{D}_{base}$ (the QSTs of $X_{base}$ are denoted as $Q^{i}_{base}$), and use $\mathcal{L}_{inner}$ for the inner-loop update:
\begin{equation}
\label{eq:meta_train_loss}
\mathcal{L}_{inner} = \mathcal{L}_{qac} (X_{base}, Y_{base}, g(Q^{i}_{base}; \{ \hat{\theta}_{bnb},\hat{\theta}_{rem},\theta_{meta}\}) ),
\end{equation}
\begin{equation}
\label{eq:inner}
\theta_{meta}^{inner}=\theta_{meta}-\beta \nabla_{\theta_{meta}} \mathcal{L}_{inner},
\end{equation}
where $\beta$ denotes the learning rate of the inner-loop. Secondly, we design a meta-testing step to enforce the learning on $\mathcal{D}_{base}$ to further exhibit certain properties on images with simulated unseen QSTs, $\mathcal{D}_{meta}$. Specifically, we quantify the performance of $\theta_{inner}$ on simulated unseen meta-test $\mathcal{D}_{meta}$:
\begin{equation}
\label{eq:meta_test_loss}
\mathcal{L}_{out}=\mathcal{L}_{qac}(X_{meta},Y_{meta}, g(\cdot; \{ \hat{\theta}_{bnb},\hat{\theta}_{rem},\theta^{inner}_{meta}\}) ).
\end{equation}
This meta-objective is computed with the inner-loop updated parameters $\theta_{meta}^{inner}$, but optimized based the original parameters $\theta_{meta}$. In other words, the network could learn how to generalize on the images with unseen QSTs through such a training scheme. Meanwhile, we utilize $\mathcal{L}_{basis}$ to ensure that the meta-learner $\theta_{meta}$ matches the ground truth on $\mathcal{D}_{base}$:
\begin{equation}
\label{eq:basis_loss}
\mathcal{L}_{basis} = \sum_{k}{|m(Q^{k}_{base};\theta_{meta}) - \boldsymbol{w}|},
\end{equation}
where $\boldsymbol{w}$ is the one hot encoding of $k$; $m(Q_{base}^k$;$\theta_{meta})$ denotes the quantization steps dependent feature generated by the meta-learner who takes $Q^{k}_{base}$ as input. Finally, we combine $\mathcal{L}_{inner},\mathcal{L}_{out}$, and $\mathcal{L}_{basis}$ to update $\theta_{meta}$:
\begin{equation}
\label{eq:final_opt}
\theta_{meta} \leftarrow \theta_{meta}-\gamma \nabla_{\theta_{meta}}(\mathcal{L}_{inner}+ \mathcal{L}_{out} + \mathcal{L}_{basis}).
\end{equation}

In summary, there are three types of QST: QST bases, simulated unseen QSTs (QSTs of images in $\mathcal{D}_{meta}$), and really unseen QSTs. There are also three steps: the base training, the meta-learner training (containing two sub-steps, inner-loop and out-loop), and testing. Table \ref{tab:q_data} shows which QSTs and equations are used in each step, and Table \ref{tab:q_data_2} shows which QSTs and equations are used in each sub-step of the meta-learner training. The complete training is summarized in Algorithm \ref{algorithm1}.
\begin{algorithm}
	\small
	\caption{QAM Training.}
	\label{algorithm1}
	\begin{algorithmic}[1]
		\State {\bf Input:} Dataset $\mathcal{D}=<\mathfrak{C}(x,Q),y>, Q \in Q_{\mathcal{D}}$, hyperparameters $\alpha, \beta, \gamma, ite_1, ite_2$
		\State {\bf Output:} The parameters of QAM: $\hat{\theta_{g}}=(\hat{\theta}_{bnb},\hat{\theta}_{rem},\hat{\theta}_{meta})$
		\State Statistic the QSTs in the trainset
		\State Select QST bases from $Q_{\mathcal{D}}$
		\State Sample $\mathcal{D}_{base}$ from $\mathcal{D}$
		\Statex {\{Base Training\}}
		\For {$ite$ in $ite_1$}
		\For {mini-batch $X_A, Y_A$ in $\mathcal{D}_{base}$}
        \State Compute $\mathcal{L}_{qac}$ by Equation (\ref{eq:base_train1})
        \State $\theta_{bnb},\theta_{rem} \leftarrow \theta_{bnb} - \alpha \nabla_{\theta_{bnb}} \mathcal{L}_{qac},\theta_{rem} - \alpha \nabla_{\theta_{rem}} \mathcal{L}_{qac}$
		\EndFor
		\EndFor
		\State $\hat{\theta}_{bnb},\hat{\theta}_{rem}, \leftarrow \theta_{bnb},\theta_{rem}$
		\Statex {\{Meta-learner training\}}
		\For {$ite$ in $ite_2$}
		\State Sample a mini-batch $X_B, Y_B$ from $\mathcal{D}_{base}$
		\State Compute $\mathcal{L}_{inner}$ by Equation (\ref{eq:meta_train_loss})
		\State Compute $\theta^{inner}_{meta}$ by Equation (\ref{eq:inner})
		\State Compute $\mathcal{L}_{basis}$ by Equation (\ref{eq:basis_loss})
		\State Sample a mini-batch $X_{C}, Y_{C}$ from $\mathcal{D}_{meta} = \complement_\mathcal{D}{\mathcal{D}_{base}}$
		\State Compute $\mathcal{L}_{out}$ by Equation (\ref{eq:meta_test_loss})
		\State Update $\theta_{meta}$ by Equation (\ref{eq:final_opt})
		\EndFor
		\State $\hat{\theta}_{meta} \leftarrow \theta_{meta}$
	\end{algorithmic}  
\end{algorithm}  

\begin{table}[tb!]
\begin{center}
	\caption{QSTs and equations used in the base training, the meta-learner training, and testing}
	\label{tab:q_data}
	\small
	\begin{adjustbox}{max width=\linewidth}
	\begin{tabular}{ccc}
    \toprule
		Step                     & QST        & Equation                    \\
		\midrule
		Base training                    & QST bases      &  (\ref{eq:base_train1})               \\
		\midrule
		\multirow{2}{*}{Meta-learner training} & QST bases      &  (\ref{eq:meta_train_loss}), (\ref{eq:inner}), (\ref{eq:meta_test_loss}), \\
		& Simulated Unseen QSTs &  (\ref{eq:basis_loss}), (\ref{eq:final_opt})         \\
		\midrule
		\multirow{3}{*}{Testing}    & QST bases      &    \multirow{3}{*}{$\backslash$}            \\
		& Simulated Unseen QSTs &                       \\
		& Really Unseen QSTs &                      \\
    \bottomrule
	\end{tabular}
	\end{adjustbox}
	\vspace{-3mm}
\end{center}
\end{table}
\begin{table}[tb!]
	\begin{center}
	\caption{QSTs and equations used in the two sub-steps of\\ the meta-learner training.}
	\small
	\label{tab:q_data_2}
    \begin{adjustbox}{max width=\linewidth}
	\begin{tabular}{cccc}
    \toprule
		Step                     & Sub-step                 & QST      & Equation                                      \\
		\midrule
		\multirow{3}{*}{Meta-learner training} & inner-loop                  & QST bases      &   (\ref{eq:meta_train_loss}), (\ref{eq:inner}) \\
		\cline{2-4}
		& \multirow{2}{*}{out-loop} & QST bases      &     \multirow{2}{*}{(\ref{eq:basis_loss}),  (\ref{eq:meta_test_loss}), (\ref{eq:final_opt})}       \\
		&                      & Simulated Unseen QSTs &   \\
    \bottomrule
	\end{tabular}
	\end{adjustbox}
	\vspace{-3mm}
	\end{center}
\end{table}
\begin{table}[!tb]
\begin{center}
	\caption{QSTs in CIFAR-10-J and CIFAR-100-J.}
    \small
	\label{tab:statistic cifar}
	\begin{adjustbox}{max width=\linewidth}
	\begin{tabular}{cc}
    \toprule
     &  QST\\
    \midrule
    Trainset & $Q_{90},Q_{80},Q_{70},Q_{60},Q_{50},Q_{40},Q_{30},Q_{20},Q_{10}$       \\
    \multirow{2}{*}{Testset} &     $Q_{90},Q_{85},Q_{80},Q_{75},Q_{70},Q_{65},Q_{60},Q_{55},$\\
                                &  $Q_{50},Q_{45},Q_{40},Q_{35},Q_{30},Q_{25},Q_{20},Q_{15},Q_{10}$  \\ 
    \bottomrule
	\end{tabular}
	\end{adjustbox}
	\vspace{-4mm}
\end{center}
\end{table}

\section{Experiment}
\label{sec:exp} 
\subsection{Datasets}
\subsubsection{CIFAR}
We choose CIFAR-10 and CIFAR-100 \cite{cifar} as source datasets. They both contain 60,000 32$\times$32$\times$3 images, of which 50,000 are for training and 10,000 for testing.
CIFAR-10 has 10 classes, and CIFAR-100 contains 100. To simulate datasets that contain different QSTs. We utilize the default quantization tables and adjust the quality factor (QF) to produce different QSTs. For simplicity, we denote the QST corresponding to $\mathrm{QF}=qf$ as $Q_{qf}$. We compress each image in source datasets by JPEG using 9 QFs (90, 80, 70, 60, 50, 40, 30, 20, 10). Moreover, we compress each testing image in source datasets to measure the model's resilience to images with unseen QSTs using an additional 9 QFs (95, 85, 75, 65, 55, 45, 35, 25, 15). As shown in Table \ref{tab:statistic cifar}, we finally conduct a dataset containing 450,000 training images and 180,000 testing images, named CIFAR-10-J and CIFAR-100-J. In some actual applications, images are not compressed heavily. Therefore, we evaluate QAM under another setting: compressing each training image in CIFAR-10 and CIFAR-100 by JPEG using 6 QFs (95, 90, 85, 80, 75, 70). Moreover, we compress each testing image using 30 QFs (99, 98, 97, ..., 70). Finally, we conduct a dataset containing 30,000 training images and 300,000 testing images, named CIFAR-10-J-S and CIFAR-100-J-S.
\begin{figure}[t!]
    \centering
    \includegraphics[width=0.8\linewidth]{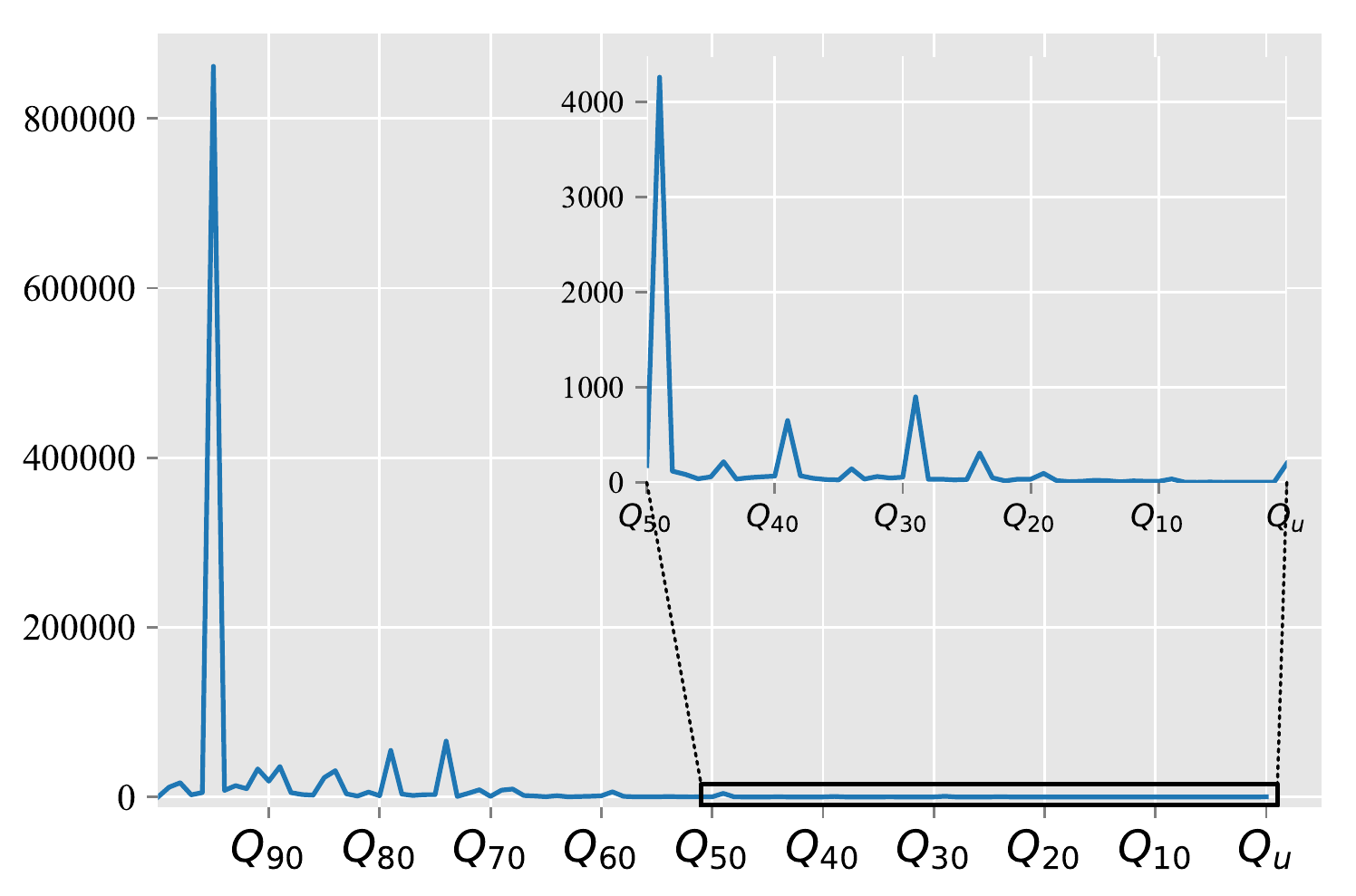}
    \vspace{-2mm}
    \caption{The imbalanced distribution of QSTs in ImageNet trainset. $Q_{u}$ denotes the QSTs that are not produced by the default quantization tables.}
    \label{fig:im_dist}
    \vspace{-3mm}
\end{figure}
\begin{figure*}[!t]
\centering
\subfloat[]{
    \begin{minipage}[b]{0.38\linewidth}
    \includegraphics[width=1\linewidth]{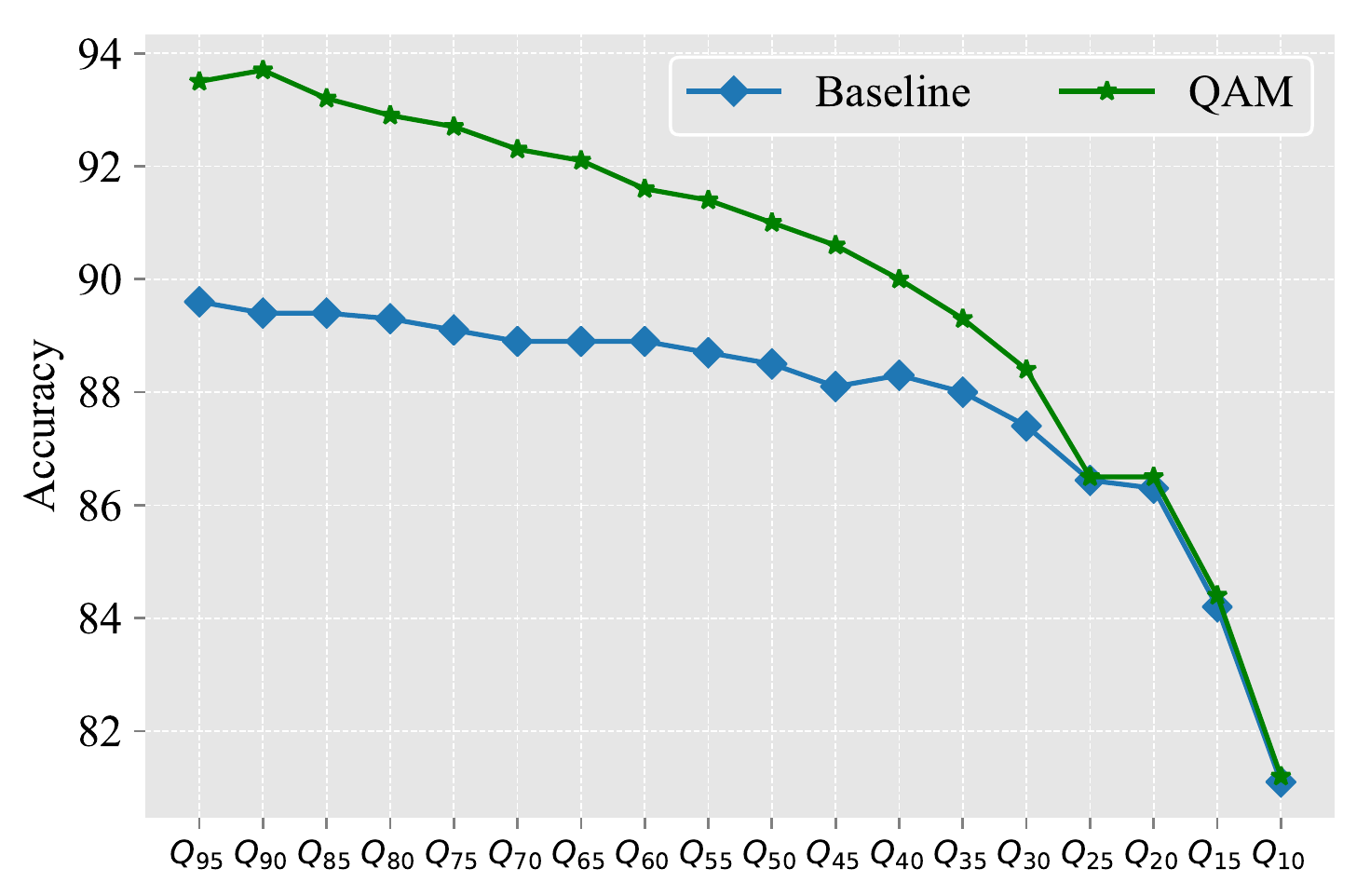}
    \end{minipage}
}
\subfloat[]{
    \begin{minipage}[b]{0.38\linewidth}
    \includegraphics[width=1\linewidth]{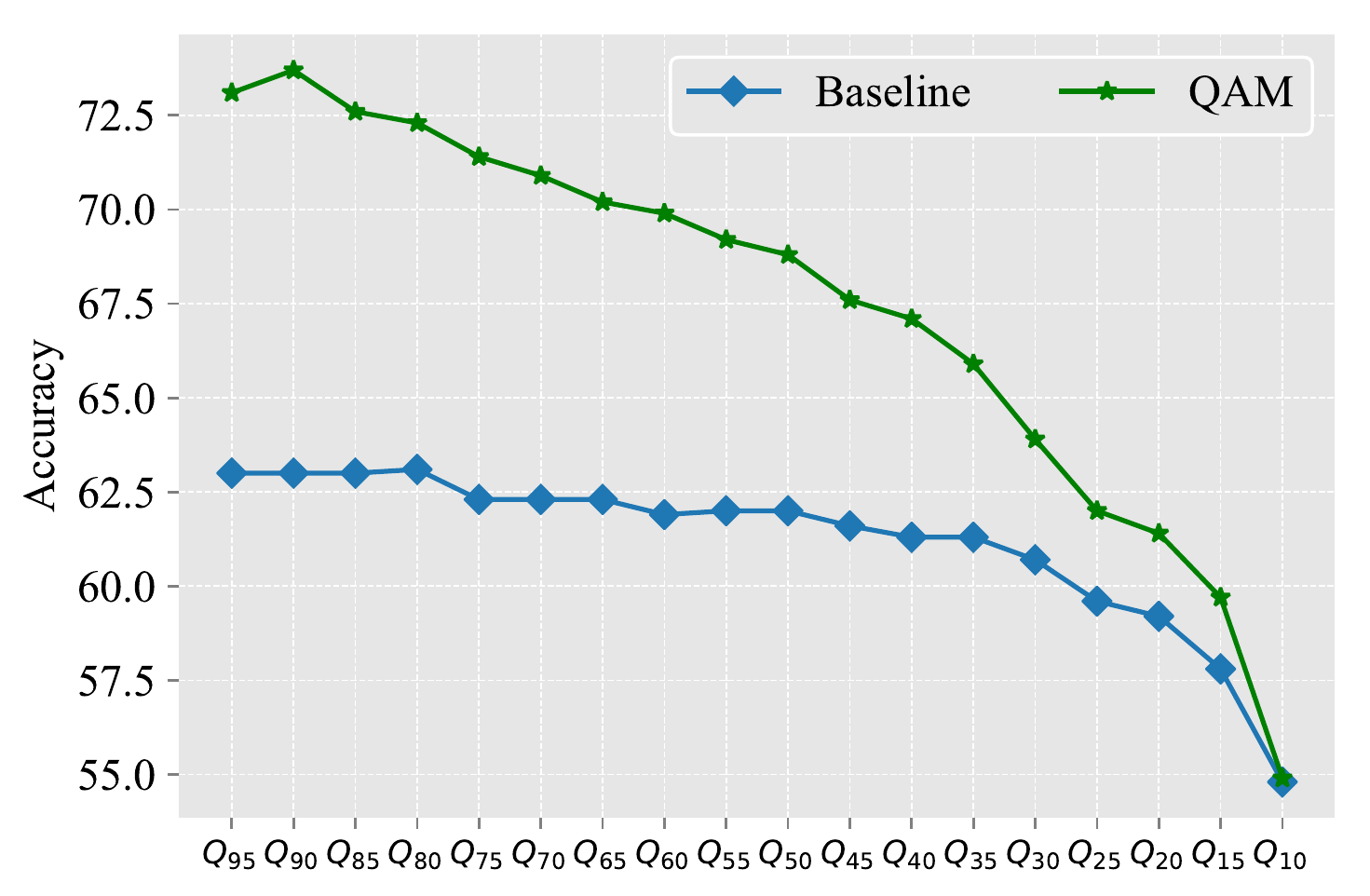}
    \end{minipage}
}
\caption{The performance of ResNet-18 for different quantization tables on (a) CIFAR-10-J and (b) CIFAR-100-J.}
\label{fig:cifar-j}
\vspace{-4mm}
\end{figure*}
\subsubsection{ImageNet}
ImageNet \cite{deng2009imagenet} is an already compressed dataset, which contains 1,000 classes of approximately 1.2 million training images and 50,000 testing images. We statistic the QSTs of images in ImageNet. As Fig. \ref{fig:im_dist} shows, ImageNet is an unbalanced dataset. The images with $Q_{96}$ account for nearly $80\%$. To measure the resilience to different QSTs, we conduct ImageNet-J, which contains 18 QSTs. Specifically, we compress each testing image using 18 QFs (95, 90, 85, 80, 75, 70, 65, 60, 55, 50, 45, 40, 35, 30, 25, 20, 15, 10). 
\subsection{CIFAR Classification}
\subsubsection{Training Step}
CIFAR-10-J and CIFAR-100-J are balanced datasets where the numbers of images with different QST are equal. We set the number of QST bases equal to the meta to balance the base training and meta training. Specifically, We choose $Q_{90}$, $Q_{60}$, $Q_{30}$, $Q_{10}$ as QST bases for CIFAR-10/100-J, and $Q_{90}$, $Q_{80}$, $Q_{70}$ for CIFAR-10/100-J-S. On the other hand, different ways of dividing is discussed in ablation studies. Based on QST bases, training images are separated into $\mathcal{D}_{base}$ and $\mathcal{D}_{meta}$. We adopt ResNet-18 \cite{He2016Deepresnet} as the baseline model. All settings and configurations in the base training of QAM are the same as those in the baseline model training. Specifically, SGD with Nesterov momentum is utilized as optimizer to update $\theta_{bnb}$ and $\theta_{rem}$, and the learning rate $\alpha$ is set to 0.1, weight decay to 0.0005, dampening to 0, momentum to 0.9. The learning rate dropped by 0.2 at 60, 120, and 160 epochs, the base training epochs $ite_1$ are set to 200, and the minibatch size is set to 128.  All input images are padded by 4 pixels on each side with reflections of the original image and pre-processed with horizontal flips and random crops. In the meta-learner training, we use Adam\cite{kingma2014adam} to update $\theta_{meta}$, and the meta-optimization step size $\beta$ is fixed to 0.001, the meta-learning rate $\gamma$ to 0.004, the meta-learner training epochs $ite_2$ to 150. 

\subsubsection{Results on CIFAR-10/100-J and CIFAR-10/100-J-S}
\begin{figure*}[th!]
\centering
\subfloat[]{
    \begin{minipage}[b]{0.38\linewidth}
    \includegraphics[width=1\linewidth]{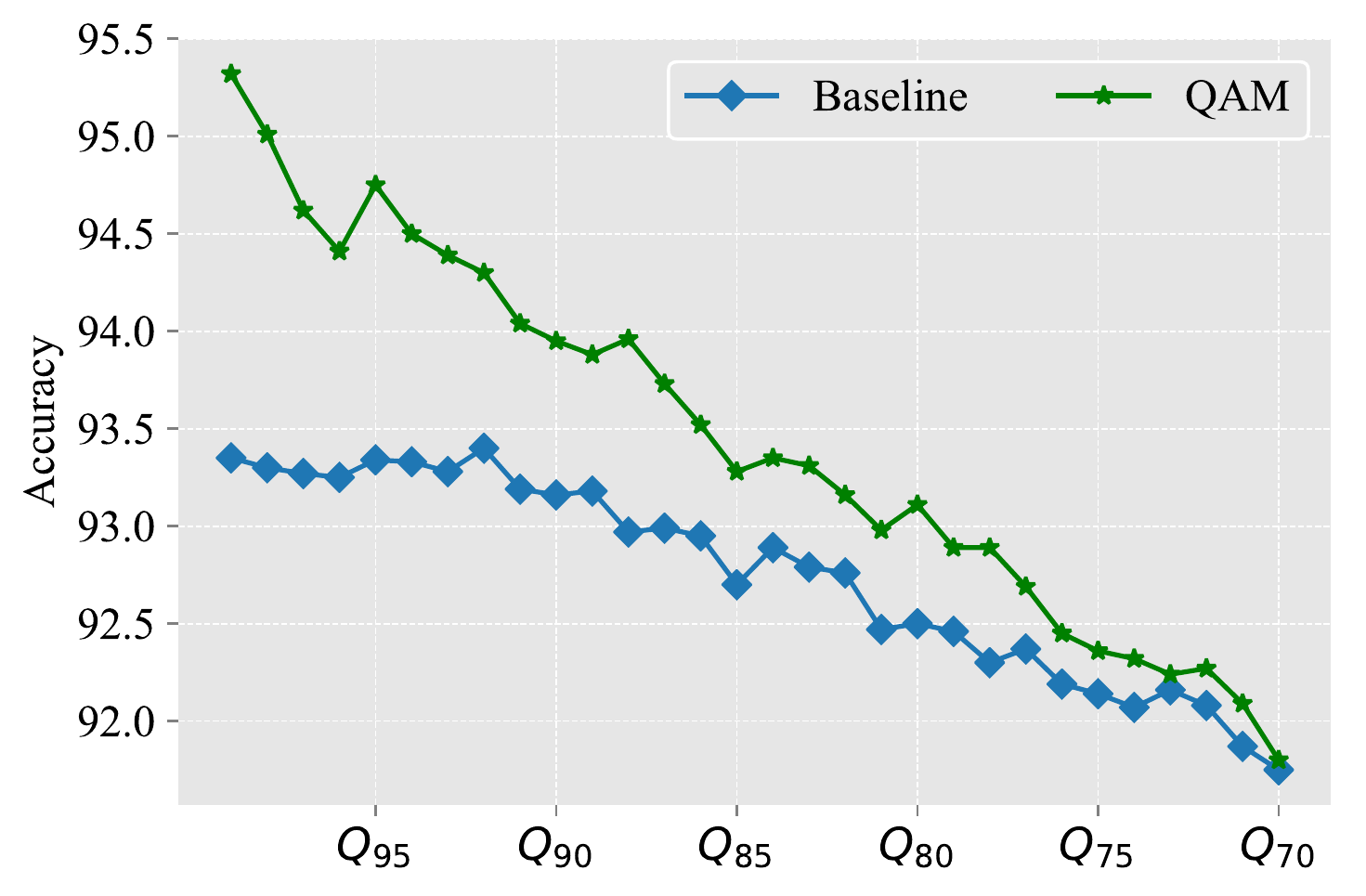}
    \end{minipage}
}
\subfloat[]{
    \begin{minipage}[b]{0.38\linewidth}
    \includegraphics[width=1\linewidth]{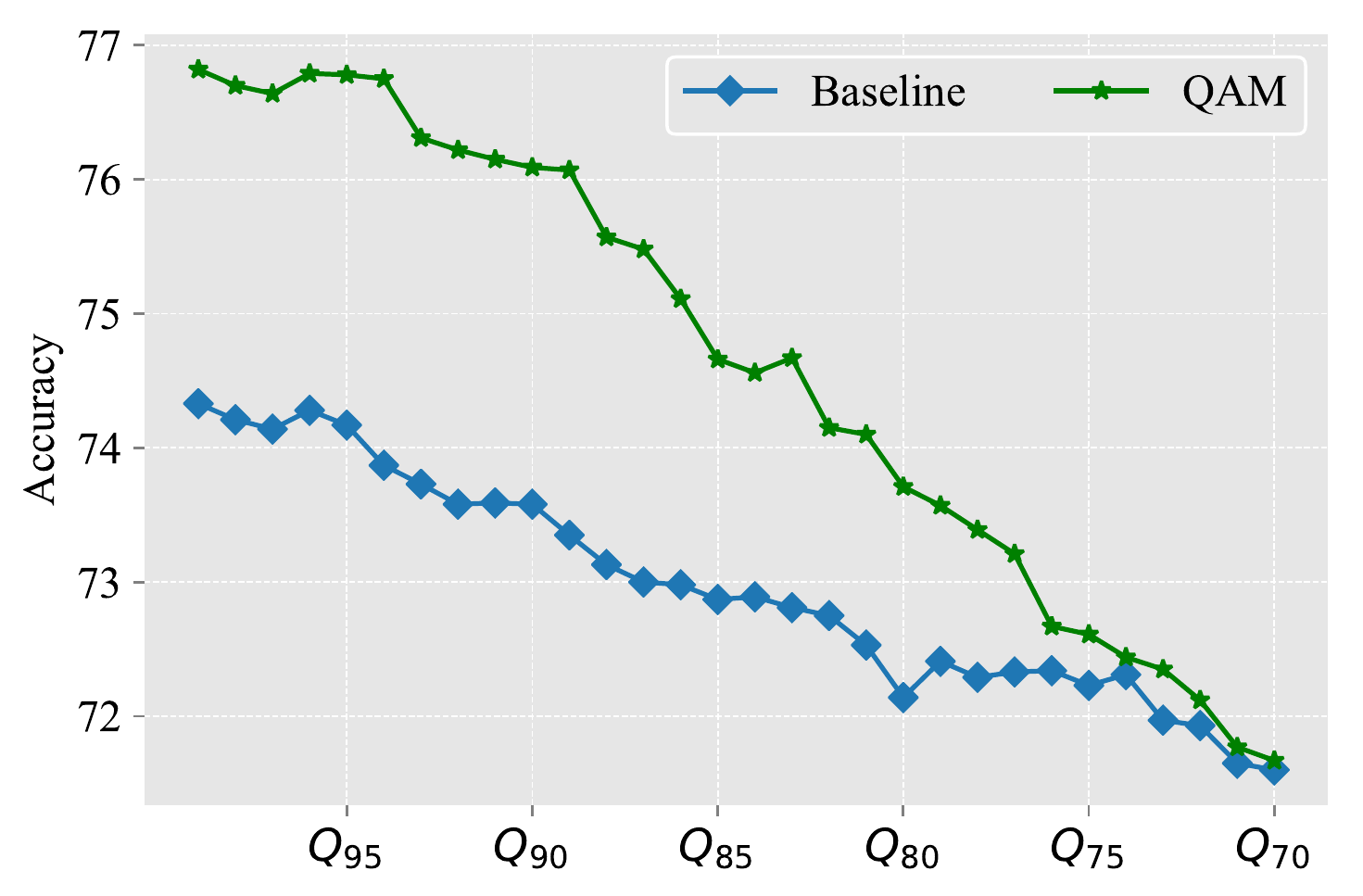}
    \end{minipage}
}
\caption{The performance of ResNet-18 for different QSTs on (a) CIFAR-10-J-S and (b) CIFAR-100-J-S.}
\label{fig:cifar-j-100-70}
\vspace{-3mm}
\end{figure*}

We compare QAM to the baseline model, which trains ResNet-18 on all training images of CIFAR-10/100-J or CIFAR-10/100-J-S and updates the parameters depending on Equation (\ref{eq:baseline}). From the results shown in Fig. \ref{fig:cifar-j} and Fig. \ref{fig:cifar-j-100-70}, we obtain three observations. Firstly, QAM achieves significant improvements, demonstrating its effectiveness. Secondly, the improvements on images with slighter compression loss are larger. The reason is that the baseline model is badly misled by heavily compressed images, while QAM utilizes QAC to reduce the interference of quantization. Thirdly, QAM does not undermine the performance over images with heavy compression but achieves slight improvements.  
\begin{figure}[th]
\vspace{-3mm}
\centering
\subfloat[]{
    \begin{minipage}[b]{0.39\linewidth}
    \includegraphics[width=1\linewidth]{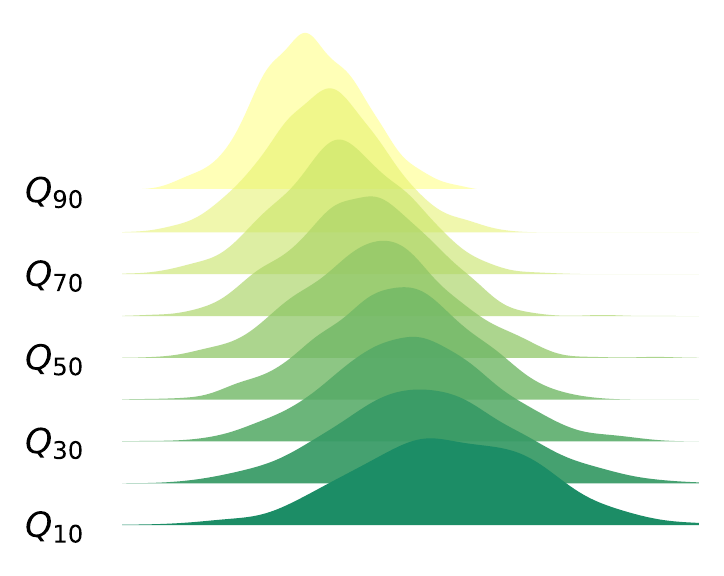} \\
    \includegraphics[width=0.9\linewidth]{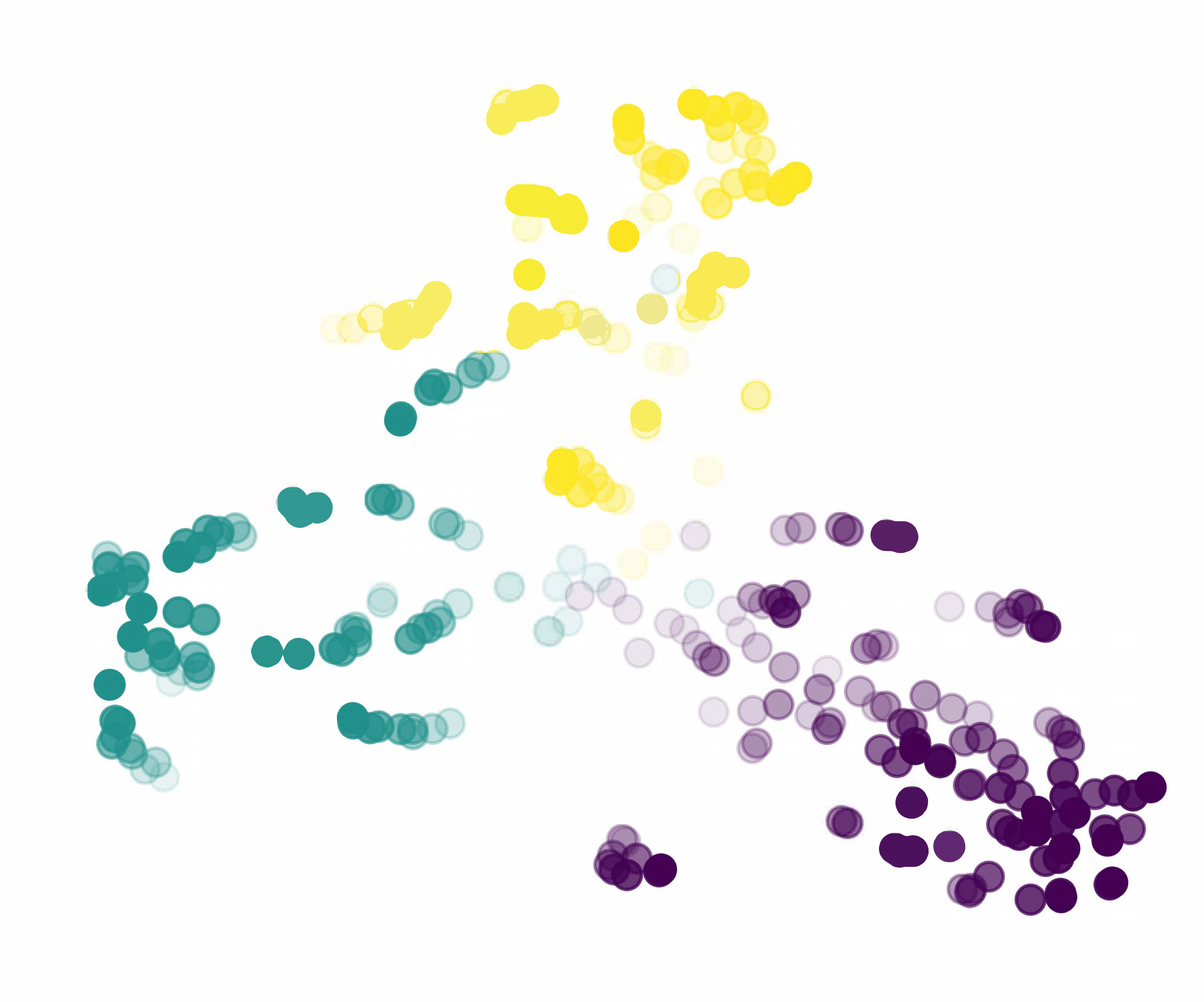} 
    \end{minipage}
}
\subfloat[]{
    \begin{minipage}[b]{0.39\linewidth}
    \includegraphics[width=1\linewidth]{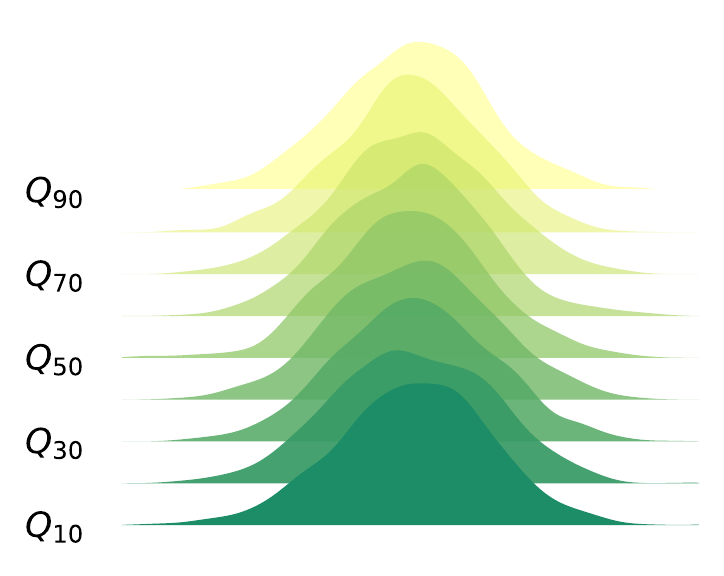} \\
    \includegraphics[width=0.9\linewidth]{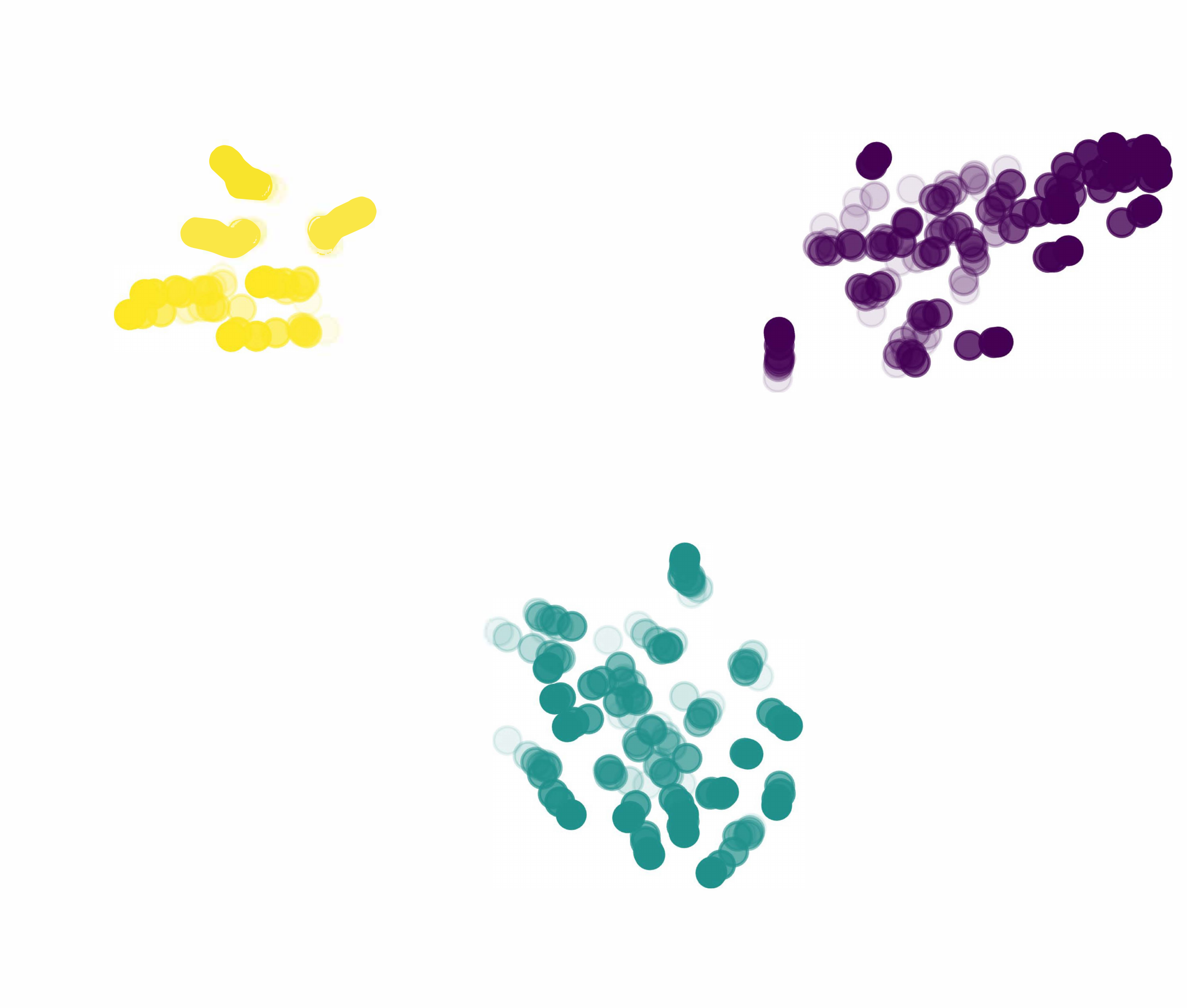}
    \end{minipage}
}
\caption{Distributions of features for (a) baseline and (b) QAM. The first row represents the distributions of the outputs of the last residual block in ResNet-18, and the second row illustrates the t-SNE \cite{maaten2008visualizing} plots of embedding features. For t-SNE plots, the lower QF, the lighter the color.}
\label{fig:feature_d}
\vspace{-4mm}
\end{figure}

\begin{table}[t!]
\caption{
Comparison with ensemble learning-based and feature alignment-based methods on CIFAR-10-J.}
\centering
\small
\label{tab:sota_el_fa}
\begin{adjustbox}{max width=\linewidth}
\begin{tabular}{cccccc}
\toprule
\multirow{2}*{Methods} & \multicolumn{2}{c}{Ensemble Learning} & \multicolumn{2}{c}{Feature Alignment} &   \multirow{2}*{QAM (ours)}            \\
\cmidrule{2-5}
 & CDI\cite{endo2020classifying}               & ADL\cite{endo2021cnnbased}               & FCT\cite{wan2020featureconsistency}                & ARF\cite{ma2021reducing}              &     \\
\midrule
$Q_{90}$                      & 90.2              & 90.0                & $\backslash$    & $\backslash$     & \textbf{93.7} \\
$Q_{80}$                & 88.8              & 89.0                & $\backslash$       & $\backslash$   & \textbf{92.9} \\
$Q_{70}$                & 88.5              & 88.8              & $\backslash$       & $\backslash$   & \textbf{92.3} \\
$Q_{60}$                & 87.6              & 88.2              & $\backslash$       & $\backslash$   & \textbf{91.6} \\
$Q_{50}$                & 87.1              & 87.8              & $\backslash$       & $\backslash$    & \textbf{91.0}   \\
$Q_{40}$                & 86.3              & 87.1              & 77.0              & 88.0               & \textbf{90.0}   \\
$Q_{30}$                & 85.8              & 86.3              & 76.3              & 86.6             & \textbf{88.4} \\
$Q_{20}$                & 83.2              & 83.8              & 74.8              & 84.2             & \textbf{86.5} \\
$Q_{10}$                & 77.7              & 78.5              & 70.3              & 78.1             & \textbf{81.2} \\
\bottomrule
\end{tabular}
\end{adjustbox}
\vspace{-3mm}
\end{table}

\begin{table}[!tb]
\caption{Ablation Studies for QAC and QABN (containing parallel BNs and meta-learner). Avg. (S) denotes average performance on QSTs in trainset; while Avg. (U) denotes average performance on QSTs that are unseen in training. All values are percentages, and the best results are indicated in bold.}
\centering
\small
\label{tab:abs1}
\begin{adjustbox}{max width=\linewidth}
\begin{tabular}{cccccc}
\toprule
QAC & Parallel BNs & Meta-learner & Avg.(S)       & Avg.(U)       & Avg.          \\
\midrule
\XSolidBrush   &\XSolidBrush & \XSolidBrush           & 64.9          & 65.9          & 65.4          \\
\CheckmarkBold  & \XSolidBrush  & \XSolidBrush       & 65.3          & 66.2          & 65.8          \\
\CheckmarkBold  & \CheckmarkBold   & \XSolidBrush       & 66.3          & 66.7          & 66.5          \\
\CheckmarkBold   & \CheckmarkBold & \CheckmarkBold            & \textbf{67.0} & \textbf{68.0} & \textbf{67.5}\\
\bottomrule
\end{tabular}
\end{adjustbox}
\end{table}

\begin{table}[!tb]
\caption{Ablation Studies for $\mathcal{L}_{inner}$, $\mathcal{L}_{out}$, and $\mathcal{L}_{basis}$. Avg.(S) denotes\\ the average performance on QSTs in trainset; while\\ Avg.(U) denotes the average performance on QSTs that are unseen in training.}
\centering
\small
\label{tab:abs2}
\begin{adjustbox}{max width=\linewidth}
\begin{tabular}{cccccc}
\toprule
$\mathcal{L}_{inner}$ & $\mathcal{L}_{out}$ & $\mathcal{L}_{basis}$ & Avg.(S)       & Avg.(U)       & Avg.          \\
\midrule
\CheckmarkBold    & \XSolidBrush & \XSolidBrush & 65.3          & 66.5          & 65.9          \\
\CheckmarkBold    & \CheckmarkBold &\XSolidBrush & 66.4          & \textbf{68.0}          & 67.2          \\
\CheckmarkBold    & \CheckmarkBold &  \CheckmarkBold & \textbf{67.0} & \textbf{68.0} & \textbf{67.5} \\
\bottomrule
\end{tabular}
\end{adjustbox}
\end{table}

Moreover, we compare QAM with several state-of-the-art methods, including two ensemble learning based methods (CDI \cite{endo2020classifying} and ADL \cite{endo2021cnnbased}) and two features alignment-based methods (FCT \cite{wan2020featureconsistency} and ARF \cite{ma2021reducing}). Table \ref{tab:sota_el_fa} report the comparison results. Although without the need for original images, QAM achieves the highest accuracy consistently.

To better understand the effects of QAM, we conduct two visualization experiments: (1) feature distributions of images with various QSTs; (2) t-SNE \cite{maaten2008visualizing} plot of embedding features. Firstly, we randomly select 90,000 testing images and draw the ridges maps for baseline and QAM. As shown in Fig. \ref{fig:feature_d}, the baseline model leads to a gap between images with different QSTs. The distances increase with increasing compression loss, indicating that the baseline model is sensitive to compression. In QAM, the distribution gaps among images with different QSTs are significantly reduced. This suggests that QAM is able to align the compression shift and lead the model to learn compression-invariant features. Secondly, we randomly select three classes from CIFAR-100-J and visualize their features with t-SNE. As shown in Fig. \ref{fig:feature_d}, the baseline has an overlap between images with different QSTs and produces more confusion for images containing heavier compression loss. By comparison, QAM could distinguish different categories clearly under images compressed at various levels.

\begin{table*}[!th]
\caption{Comparison with state-of-the-art domain generalization methods on CIFAR-10-J.}
\centering
\small
\label{tab:cifar-10-dg}
\begin{adjustbox}{max width=\linewidth}
\begin{tabular}{cccccccccccccccccccc}
\toprule
Methods & $Q_{95}$& $Q_{90}$& $Q_{85}$& $Q_{80}$& $Q_{75}$& $Q_{70}$& $Q_{65}$& $Q_{60}$& $Q_{55}$& $Q_{50}$& $Q_{45}$& $Q_{40}$& $Q_{35}$& $Q_{30}$& $Q_{25}$& $Q_{20}$& $Q_{15}$& $Q_{10}$& Avg. \\
\midrule
SD \cite{SDpezeshki2020gradient}      & 89.7          & 89.7          & 89.5          & 89.6          & 89.5          & 89.4          & 89.2          & 89.1          & 88.7          & 88.6          & 88.5          & 88.5          & 88.1          & 87.6          & 86.7          & 85.9          & 84.1          & \textbf{82.0} & 87.9          \\
SagNet \cite{SagNetnam2021reducing}  & 89.9          & 89.8          & 89.8          & 89.6          & 89.4          & 89.4          & 89.5          & 89.2          & 88.9          & 89.0          & 88.8          & 88.7          & 88.5          & 88.2          & 86.9          & 86.1          & 84.8          & \textbf{82.0} & 88.1          \\
SelfReg \cite{kim2021selfreg} & 89.8          & 89.7          & 89.7          & 89.6          & 89.6          & 89.2          & 89.3          & 89.2          & 88.9          & 88.9          & 88.6          & 88.7          & 88.6          & 87.9          & 86.9          & 86.2          & 84.8          & 80.8          & 88.0          \\
MLDG \cite{mldgli2018learning}    & 89.9          & 89.7          & 89.7          & 89.6          & 89.4          & 89.4          & 89.4          & 89.1          & 88.8          & 88.8          & 88.6          & 88.6          & 88.3          & 87.9          & 86.8          & 86.0          & 84.5          & \textbf{82.0} & 88.0          \\
MMD \cite{mmdli2018domain}     & 89.7          & 89.6          & 89.6          & 89.3          & 89.0          & 89.3          & 89.1          & 89.0          & 88.9          & 88.9          & 88.8          & 88.8          & 88.3          & 88.3          & 87.0          & 86.2          & 84.8          & 81.6          & 88.0          \\
CORAL \cite{coralsun2016deep}   & 90.2          & 90.0          & 89.9          & 90.0          & 89.7          & 89.5          & 89.3          & 89.4          & 89.2          & 89.1          & 89.0          & 89.1          & 88.8          & \textbf{88.4} & \textbf{87.2} & \textbf{86.5} & \textbf{84.9} & 81.8          & 88.3          \\
MTL \cite{mtlblanchard2017domain}     & 89.4          & 89.3          & 89.3          & 89.2          & 89.1          & 88.8          & 88.8          & 88.8          & 88.7          & 88.7          & 88.4          & 88.5          & 88.4          & 87.6          & 86.6          & 85.7          & 84.6          & 80.8          & 87.7          \\
Fish \cite{fishshi2021gradient}    & 90.0          & 89.9          & 89.8          & 89.7          & 89.5          & 89.5          & 89.4          & 89.3          & 89.0          & 89.0          & 88.8          & 88.8          & 88.5          & 88.2          & 87.1          & 86.3          & 84.8          & 81.8          & 88.2          \\
\midrule
QAM (ours)     & \textbf{93.5} & \textbf{93.7} & \textbf{93.2} & \textbf{92.9} & \textbf{92.7} & \textbf{92.3} & \textbf{92.1} & \textbf{91.6} & \textbf{91.4} & \textbf{91.0} & \textbf{90.6} & \textbf{90.0} & \textbf{89.3} & \textbf{88.4} & 86.5          & \textbf{86.5} & 84.4          & 81.2          & \textbf{90.1} \\
\bottomrule
\end{tabular}
\end{adjustbox}
\vspace{-2mm}
\end{table*}

\begin{table*}[!tb]
\caption{Comparison with state-of-the-art domain generalization methods on CIFAR-100-J.}
\centering
\small
\label{tab:cifar-100-dg}
\begin{adjustbox}{max width=\linewidth}
\begin{tabular}{cccccccccccccccccccc}
\toprule
Methods & $Q_{95}$& $Q_{90}$& $Q_{85}$& $Q_{80}$& $Q_{75}$& $Q_{70}$& $Q_{65}$& $Q_{60}$& $Q_{55}$& $Q_{50}$& $Q_{45}$& $Q_{40}$& $Q_{35}$& $Q_{30}$& $Q_{25}$& $Q_{20}$& $Q_{15}$& $Q_{10}$& Avg. \\
\midrule
SD \cite{SDpezeshki2020gradient}      & 62.8          & 63.0          & 62.7          & 62.7          & 62.4          & 62.2          & 62.1          & 62.0          & 61.7          & 61.7          & 61.6          & 61.5          & 61.2          & 60.8          & 60.3          & 59.1          & 57.5          & 54.1          & 61.1          \\
SagNet \cite{SagNetnam2021reducing}  & 63.2          & 62.9          & 62.9          & 63.0          & 62.9          & 62.4          & 62.4          & 62.2          & 62.0          & 61.9          & 61.8          & 61.5          & 61.3          & 60.3          & 59.8          & 59.3          & 58.0          & 53.8          & 61.2          \\
SelfReg \cite{kim2021selfreg} & 63.7          & 63.8          & 63.8          & 63.7          & 63.5          & 63.1          & 63.0          & 63.1          & 62.7          & 62.5          & 62.4          & 62.4          & 62.1          & 61.4          & 60.6          & 60.0          & 58.4          & \textbf{55.1} & 62.0          \\
MLDG \cite{mldgli2018learning}    & 60.1          & 59.9          & 59.9          & 59.9          & 60.0          & 59.2          & 59.2          & 59.7          & 58.9          & 59.1          & 58.9          & 58.8          & 58.1          & 57.6          & 57.0          & 56.3          & 55.2          & 50.8          & 58.2          \\
MMD \cite{mmdli2018domain}     & 62.9          & 63.0          & 63.0          & 63.1          & 62.8          & 62.7          & 62.5          & 62.3          & 62.0          & 62.2          & 62.0          & 61.7          & 61.7          & 61.2          & 60.8          & 59.6          & 57.6          & 54.2          & 61.4          \\
CORAL \cite{coralsun2016deep}   & 63.3          & 63.5          & 63.4          & 63.6          & 63.5          & 63.2          & 63.1          & 62.9          & 62.6          & 62.5          & 62.6          & 62.2          & 62.0          & 61.0          & 60.8          & 59.8          & 58.0          & 54.0          & 61.8          \\
MTL \cite{mtlblanchard2017domain}     & 63.3          & 63.1          & 63.0          & 62.7          & 62.8          & 63.1          & 62.7          & 62.3          & 62.1          & 62.5          & 62.3          & 61.4          & 61.6          & 61.1          & 60.5          & 59.7          & 57.5          & 53.7          & 61.4          \\
Fish \cite{fishshi2021gradient}    & 63.3          & 63.4          & 63.4          & 63.4          & 62.9          & 62.7          & 62.7          & 62.5          & 62.3          & 62.2          & 62.0          & 61.9          & 61.7          & 61.0          & 60.1          & 59.6          & 58.1          & 54.9          & 61.6          \\
\midrule
QAM (ours)    & \textbf{73.1} & \textbf{73.7} & \textbf{72.6} & \textbf{72.3} & \textbf{71.4} & \textbf{70.9} & \textbf{70.2} & \textbf{69.9} & \textbf{69.2} & \textbf{68.8} & \textbf{67.6} & \textbf{67.1} & \textbf{65.9} & \textbf{63.9} & \textbf{62.0} & \textbf{61.4} & \textbf{59.7} & 54.9          & \textbf{67.5}
 \\
\bottomrule
\end{tabular}
\end{adjustbox}
\vspace{-2mm}
\end{table*}
\begin{figure}[!t]
\centering
\includegraphics[width=1.00\linewidth]{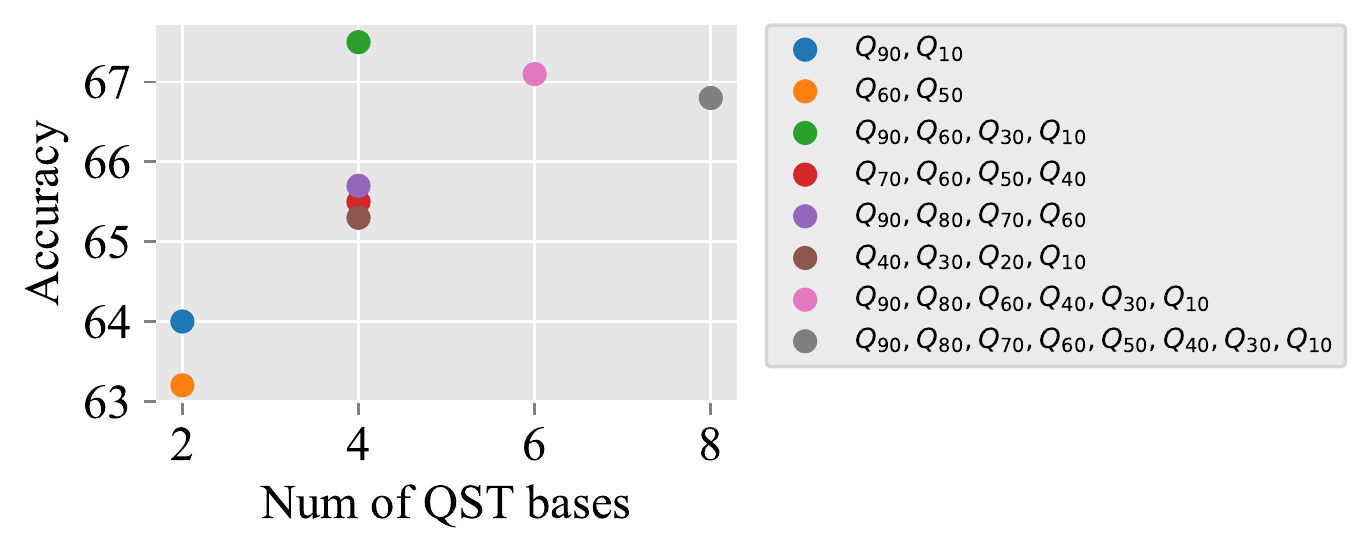}
\vspace{-3mm}
\caption{The performance of QAM when utilizing different QSTs as QST bases. Different color points denote QAM with varying QST bases.}
\label{fig:bq_cifar}
\end{figure}
\begin{figure*}[!t]
\vspace{-3mm}
\centering
\subfloat[]{
    \begin{minipage}[b]{0.38\linewidth}
    \includegraphics[width=1\linewidth]{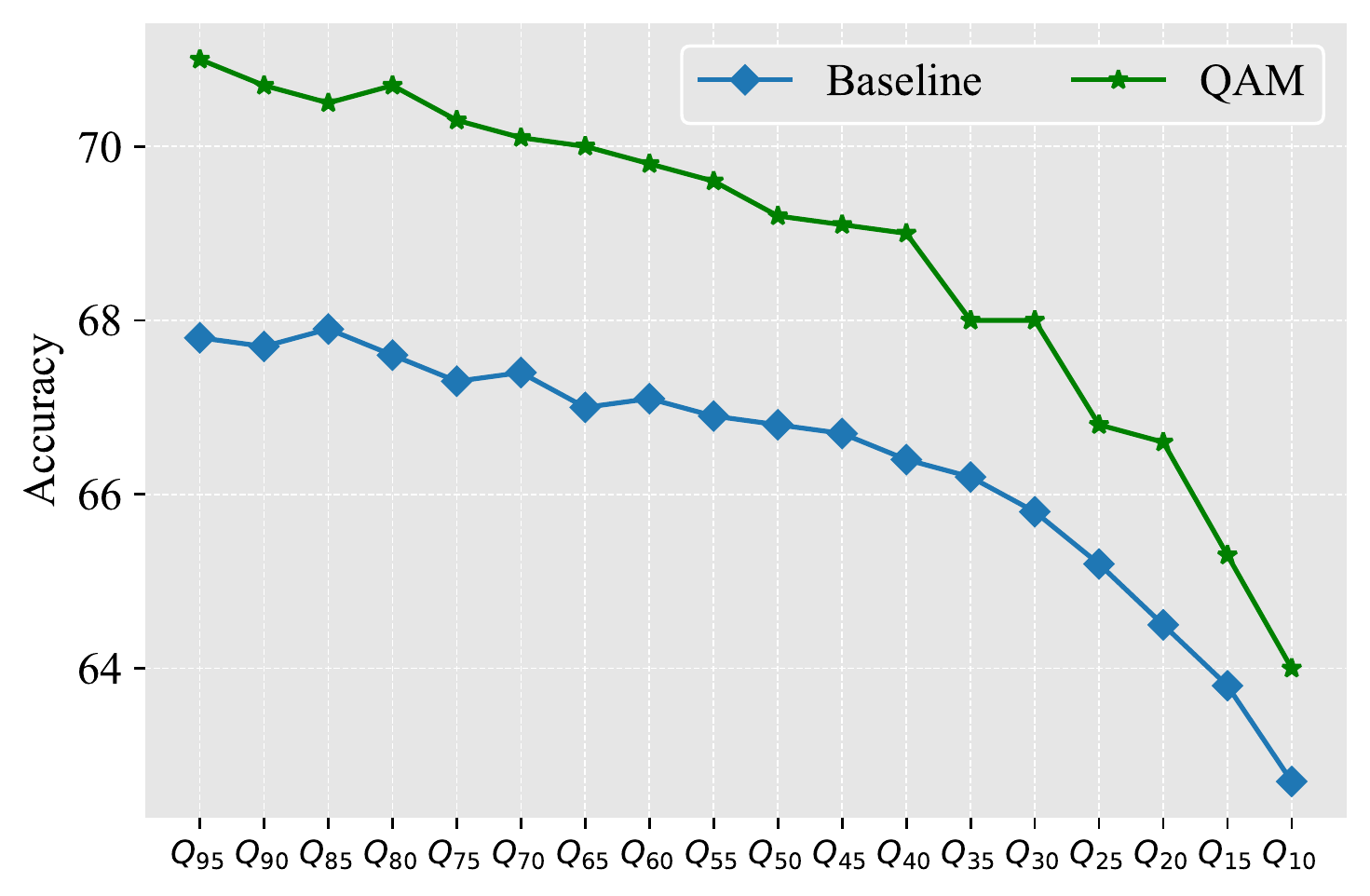}
    \end{minipage}
}
\subfloat[]{
    \begin{minipage}[b]{0.38\linewidth}
    \includegraphics[width=1\linewidth]{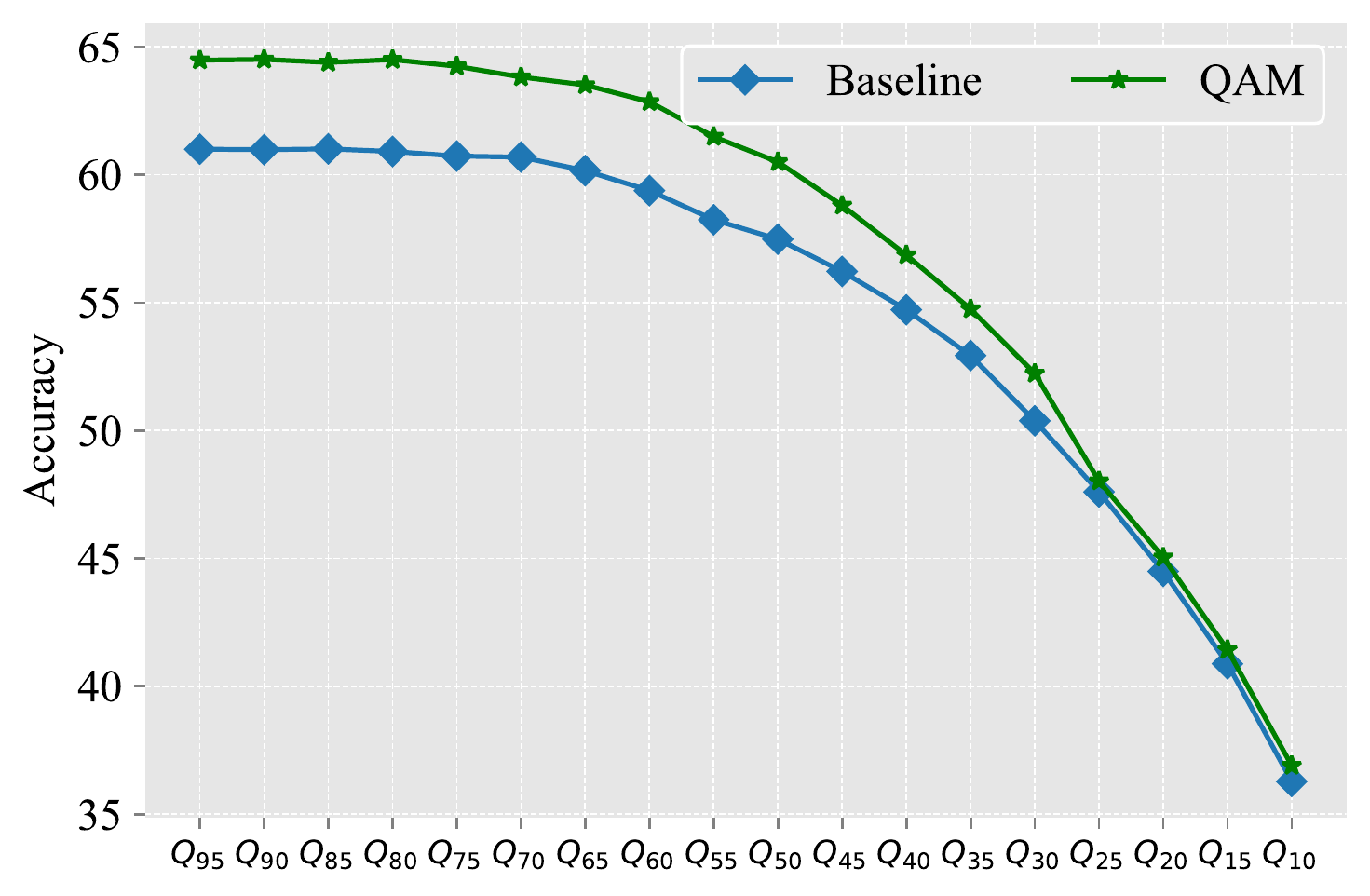}
    \end{minipage}
}
\caption{The performance of ResNet-18 on cifar-100 compressed with (a) WebP\cite{webp} and (b) JPEG2000\cite{jpeg2000christopoulos2000}.}
\label{fig:cifar-100-2k}
\end{figure*}
\subsubsection{Ablation Studies}
Firstly, we conduct experiments to investigate the effectiveness of QAC and QABN. To detailly evaluate QABN, we separate it into two parts, parallel BNs and the meta-learner. The results are reported in Table \ref{tab:abs1}. It could be observed that without using these three components, the baseline model achieves poor results $65.4\%$. When utilizing QAC as sample weights, the model could achieve $0.4\%$ improvement ($65.4\%\rightarrow 65.8\%$) than the baseline model, demonstrating the effectiveness of QAC. When replacing BNs with parallel BNs, the model could achieve $0.7\%$ improvement ($65.8\%\rightarrow 66.5\%$). Furthermore, the model with meta-learner could achieve $1.0\%$ improvement ($66.5\% \rightarrow 67.5\%$). Moreover, the improvement on QSTs that are unseen in training is greater ($1.3\%: 66.7\%\rightarrow 68.0\%$), demonstrating that the meta-learner could improve the generalization ability of QAM to unseen QSTs.

Secondly, experiments are conducted to investigate the effectiveness of the three losses to train the meta-learner, including $\mathcal{L}_{inner}$, $\mathcal{L}_{out}$, and $\mathcal{L}_{basis}$. From Table \ref{tab:abs2}, we could observe that only utilizing $\mathcal{L}_{inner}$, the model achieves poor results $65.9\%$. When adding $\mathcal{L}_{out}$, the model achieves significant improvements on unseen QSTs ($66.5\%\rightarrow 68.0\%$). When combining the three losses, the model could get further improvements both on seen and unseen QSTs.

Thirdly, experiments are conducted to investigate the influence of QST bases. CIFAR-10-J and CIFAR-100-J are balanced datasets, where the numbers of images with different QSTs are equal. We set the number of QST bases equal with the meta to balance the base training and meta-learner training (the results of different settings are shown in Fig. \ref{fig:bq_cifar}). Meanwhile, we observe that the performance on $\{Q_{90},Q_{60},Q_{30},Q_{10}\}$ is better than those on $\{Q_{90},Q_{80},Q_{70},Q_{60}\}$, $\{Q_{70},Q_{60},Q_{50},Q_{40}\}$, and $\{Q_{40},Q_{30},Q_{20},Q_{10}\}$, and the performance on $\{Q_{90},Q_{10}\}$ is better than the one on $\{Q_{60},Q_{50}\}$. The results are in line with our intuition that a wider basis diversity makes better performance.
\begin{figure*}[tb!]
\centering
\includegraphics[width=0.78\linewidth]{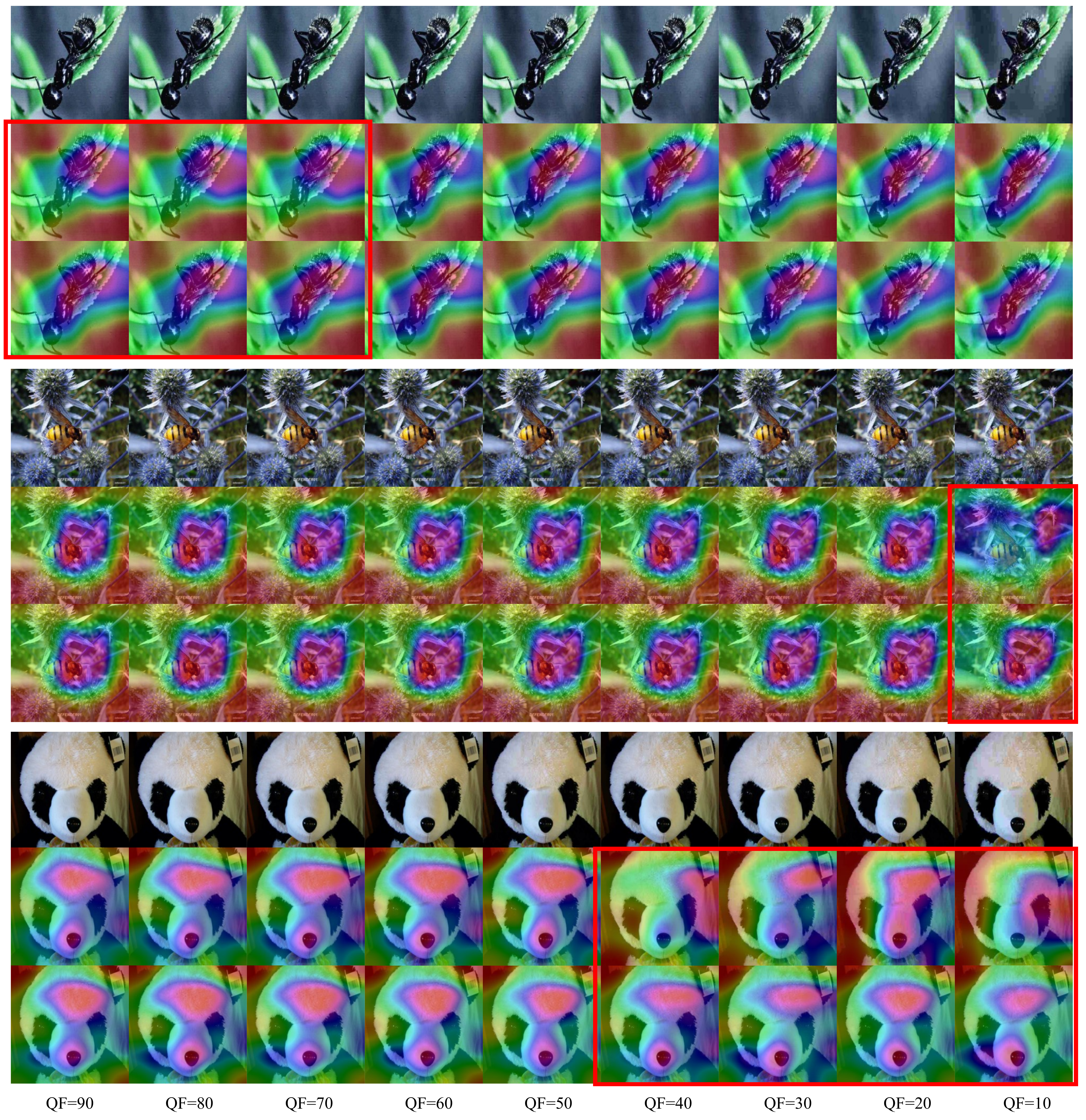}
\vspace{-2mm}
\caption{The Gradient-weighted Class Activation Mapping of the baseline (the second axis) and QAM (the third axis) for images with different QSTs (the first axis). The first sample suffers from slight compression loss, while the other samples suffer from heavy compression loss, which is marked by red boxes.}
\vspace{-1mm}
\label{fig:gradmap}
\end{figure*}
\begin{table*}[!tb]
\caption{The performance of 40-2 Wide ResNet and ResNeXt-29($32\times 4$) on CIFAR-10-J and CIFAR-100-J.}
\centering
\small
\label{tab:wide-resnet}
\begin{adjustbox}{max width=\linewidth}
\begin{tabular}{ccccccccccccccccccccc}
\toprule
\multicolumn{2}{c}{CIFAR-10-J} &$Q_{95}$& $Q_{90}$& $Q_{85}$& $Q_{80}$& $Q_{75}$& $Q_{70}$& $Q_{65}$& $Q_{60}$& $Q_{55}$& $Q_{50}$& $Q_{45}$& $Q_{40}$& $Q_{35}$& $Q_{30}$& $Q_{25}$& $Q_{20}$& $Q_{15}$& $Q_{10}$& Avg. \\
\midrule
\multirow{2}*{WRN\cite{zagoruyko2016wide}} &Baseline & 87.8          & 88.4          & 87.8          & 87.7          & 87.1          & 86.8          & 87.0          & 86.2          & 85.9          & 85.8          & 85.1          & 84.9          & 83.6          & 82.7          & 81.4          & 78.6          & 76.9          & 76.5          & 84.4          \\
&QAM      & \textbf{91.0} & \textbf{91.4} & \textbf{90.8} & \textbf{90.9} & \textbf{90.3} & \textbf{90.0} & \textbf{89.6} & \textbf{89.1} & \textbf{88.8} & \textbf{88.4} & \textbf{87.7} & \textbf{87.2} & \textbf{86.0} & \textbf{84.9} & \textbf{83.3} & \textbf{80.1} & \textbf{78.3} & \textbf{77.4} & \textbf{86.9} \\
\multirow{2}*{ResNeXt\cite{xie2017aggregated}} &Baseline & 90.8          & 90.6          & 90.5          & 90.4          & 90.3          & 90.2          & 90.2          & 90.1          & 89.6          & 89.4          & 89.1          & 88.9          & 88.4          & 87.8          & 86.8          & 86.5          & 84.3          & 80.9          & 88.6          \\
&QAM      &\textbf{93.3} & \textbf{93.3} & \textbf{92.8} & \textbf{92.7} & \textbf{92.3} & \textbf{92.2} & \textbf{91.9} & \textbf{91.3} & \textbf{90.7} & \textbf{90.4} & \textbf{89.9} & \textbf{89.5} & \textbf{88.9} & \textbf{88.4} & \textbf{87.6} & \textbf{86.9} & \textbf{85.3} & \textbf{81.6} & \textbf{89.9} \\
\midrule
\multicolumn{2}{c}{CIFAR-100-J} &$Q_{95}$& $Q_{90}$& $Q_{85}$& $Q_{80}$& $Q_{75}$& $Q_{70}$& $Q_{65}$& $Q_{60}$& $Q_{55}$& $Q_{50}$& $Q_{45}$& $Q_{40}$& $Q_{35}$& $Q_{30}$& $Q_{25}$& $Q_{20}$& $Q_{15}$& $Q_{10}$& Avg. \\
\midrule
\multirow{2}*{WRN\cite{zagoruyko2016wide}} &Baseline & 66.3          & 66.3          & 66.4          & 66.2 & 65.8 & 65.8          & 65.2          & 65.1          & 64.9          & 64.6          & 64.2          & 63.9          & 62.7          & 61.2          & 61.3          & 60.4          & 56.9          & 53.1          & 63.3          \\
&QAM      &\textbf{69.6} & \textbf{72.1} & \textbf{69.4} & \textbf{69.8} & \textbf{68.9} & \textbf{68.2} & \textbf{67.8} & \textbf{67.0} & \textbf{66.6} & \textbf{66.0} & \textbf{64.8} & \textbf{64.1} & \textbf{63.3} & \textbf{62.9} & \textbf{62.4} & \textbf{60.6} & \textbf{58.0} & \textbf{53.5} & \textbf{65.3} \\
\multirow{2}*{ResNeXt\cite{xie2017aggregated}} &Baseline & 68.3          & 68.4          & 68.3          & 67.9 & 67.5 & 67.7          & 67.2          & 67.0          & 66.6          & 66.5          & 66.3          & 66.3          & 65.6          & 64.9          & 62.0          & \textbf{59.9} & 58.3          & 55.3          & 65.2          \\
&QAM      &\textbf{74.9} & \textbf{74.7} & \textbf{74.2} & \textbf{74.0} & \textbf{73.3} & \textbf{72.9} & \textbf{72.2} & \textbf{71.6} & \textbf{71.1} & \textbf{70.5} & \textbf{69.7} & \textbf{68.5} & \textbf{67.1} & \textbf{65.3} & \textbf{62.6} &         59.2  & \textbf{59.5} & \textbf{55.4} & \textbf{68.7}
\\
\bottomrule
\end{tabular}
\end{adjustbox}
\vspace{-2mm}
\end{table*}
\subsubsection{Comparison with Domain Generalization}
The goal of domain generalization \cite{SDpezeshki2020gradient,SagNetnam2021reducing,mmdli2018domain,mldgli2018learning, coralsun2016deep,fishshi2021gradient,kim2021selfreg,mtlblanchard2017domain} is to predict well on distributions different from those seen during training. If we consider the compressed images with different QSTs (i.e., different levels of compression loss) as different domains (QSTs are regarded as domain labels), domain generalization could be a solution. We compare QAM with the recent state-of-the-art domain generalization methods, including SD \cite{SDpezeshki2020gradient}, SagNet \cite{SagNetnam2021reducing}, SelfReg \cite{kim2021selfreg}, MLDG \cite{mldgli2018learning}, MMD \cite{mmdli2018domain}, CORAL \cite{coralsun2016deep}, MTL \cite{mtlblanchard2017domain}, Fish \cite{fishshi2021gradient}. Table \ref{tab:cifar-10-dg} and Table \ref{tab:cifar-100-dg} report the results. In this experiment, the domains corresponding to $Q_{90}$, $Q_{80}$, $Q_{70}$, $Q_{60}$, $Q_{50}$, $Q_{40}$, $Q_{30}$, $Q_{20}$, $Q_{10}$ are regarded as source domains, while the others are target domains. The networks are only trained on the training images of source domains but tested on testing images of all domains. Our implementation is based on DomainBed \cite{gulrajani2020search}. As shown in Table \ref{tab:cifar-100-dg}, we could observe that QAM achieves $67.5\%$ precision, outperforming domain generalization methods. The reason for this phenomenon is that equally taking images containing different levels of compression loss makes domain generalization methods misled by the heavily compressed domains. 
\begin{table*}[!t]
\caption{The performance of ResNet-50 for different QSTs on ImageNet-J.}
\centering
\small
\label{tab:imagenet-j}
\begin{adjustbox}{max width=\linewidth}
\begin{tabular}{cccccccccccccccccccc}
\toprule
ImageNet-J &$Q_{95}$& $Q_{90}$& $Q_{85}$& $Q_{80}$& $Q_{75}$& $Q_{70}$& $Q_{65}$& $Q_{60}$& $Q_{55}$& $Q_{50}$& $Q_{45}$& $Q_{40}$& $Q_{35}$& $Q_{30}$& $Q_{25}$& $Q_{20}$& $Q_{15}$& $Q_{10}$& Avg. \\
\midrule
Baseline & 75.9          & 75.8          & 75.6          & 75.4          & 75.3          & 75.1          & 74.9          & 74.7          & 74.4          & 74.3          & 73.9          & 73.6          & 73.2          & 72.7          & 71.7          & 70.4          & 68.5          & 63.2  &  73.3      \\
QAM (ours) &\textbf{76.6} & \textbf{76.5} & \textbf{76.4} & \textbf{76.2} & \textbf{76.2} & \textbf{76.1} & \textbf{76.0} & \textbf{75.8} & \textbf{75.7} & \textbf{75.5} & \textbf{75.2} & \textbf{75.0} & \textbf{74.8} & \textbf{74.3} & \textbf{73.7} & \textbf{72.6} & \textbf{70.4} & \textbf{64.6} & \textbf{74.5}
\\
\bottomrule
\end{tabular}
\end{adjustbox}
\vspace{-2mm}
\end{table*}
\begin{table*}[!tb]
\caption{
The performance of RegNetY-16GF and EfficientNetV2-S on ImageNet-J.}
\centering
\small
\label{tab:regnet}
\begin{adjustbox}{max width=\linewidth}
\begin{tabular}{ccccccccccccccccccccc}
\toprule
\multicolumn{2}{c}{ImageNet-J} &$Q_{95}$& $Q_{90}$& $Q_{85}$& $Q_{80}$& $Q_{75}$& $Q_{70}$& $Q_{65}$& $Q_{60}$& $Q_{55}$& $Q_{50}$& $Q_{45}$& $Q_{40}$& $Q_{35}$& $Q_{30}$& $Q_{25}$& $Q_{20}$& $Q_{15}$& $Q_{10}$& Avg. \\
\midrule
\multirow{2}*{RegNetY-16GF\cite{regnetradosavovic2020designing}} & Baseline & 82.3          & 82.1          & 81.9          & 81.8          & 81.7          & 81.6          & 81.5          & 81.4          & 81.2          & 81.1          & 80.8          & 80.5          & 80.2          & 79.9          & 79.4          & 78.7          & 77.2          & 73.3          & 80.4          \\
 & QAM      & \textbf{83.4} & \textbf{83.2} & \textbf{83.0} & \textbf{82.9} & \textbf{82.7} & \textbf{82.5} & \textbf{82.2} & \textbf{82.1} & \textbf{82.0} & \textbf{81.8} & \textbf{81.7} & \textbf{81.6} & \textbf{81.4} & \textbf{81.0} & \textbf{80.6} & \textbf{79.9} & \textbf{78.6} & \textbf{75.5} & \textbf{81.4} \\
\midrule
\multirow{2}*{EfficientNetV2-S\cite{tan2021efficientnetv2}} &Baseline & 82.8          & 82.6          & 82.5          & 82.3          & 81.9          & 81.7          & 81.4          & 81.2          & 81.0          & 80.9          & 80.5          & 80.4          & 80.1          & 79.7          & 79.2          & 78.4          & 76.8          & 73.8          & 80.4          \\
&QAM      &\textbf{83.3} & \textbf{83.2} & \textbf{82.9} & \textbf{82.6} & \textbf{82.3} & \textbf{82.0} & \textbf{81.8} & \textbf{81.6} & \textbf{81.5} & \textbf{81.2} & \textbf{81.0} & \textbf{80.7} & \textbf{80.4} & \textbf{80.0} & \textbf{79.4} & \textbf{78.6} & \textbf{77.2} & \textbf{74.2} & \textbf{80.8}\\
\bottomrule
\end{tabular}
\end{adjustbox}
\vspace{-4mm}
\end{table*}
\subsubsection{Generalization Capability Evidence}
Firstly, experiments are conducted to clarify the generalization capability of QAM to other encoding formats. Similar to CIFAR-100-J, we construct two datasets based on WebP\cite{webp} and JPEG2000\cite{jpeg2000christopoulos2000}. Fig. \ref{fig:cifar-100-2k} shows the results, and all settings are the same as those of the experiments on CIFAR-100-J.
It could be observed that QAM achieves significant improvements, demonstrating the effectiveness of QAM on WebP and JPEG2000. On the other hand, QAM achieves better improvements on images with higher QFs, which is similar to the results on CIFAR-100-J. The reason is that the baseline is misled by heavily compressed images, while QAM could utilize QAC to reduce the interference of compression loss.

Secondly, we investigate the generalization capability of QAM to other baseline models. Specifically, we utilize QAM to a 40-2 Wide ResNet \cite{zagoruyko2016wide}, and a ResNeXt-29($32\times 4$) \cite{xie2017aggregated}. From the results shown in Table \ref{tab:wide-resnet}, we observe that QAM achieves significant improvements on various baseline networks. The results evidence the generalization capability of QAM to other networks. 

\subsection{ImageNet Classification}
\subsubsection{Training setup}
As Fig. \ref{fig:im_dist} shows, ImageNet is an unbalanced dataset. We select the top QSTs in the number of images as QST bases. Specifically, we choose $Q_{96}$, $Q_{75}$, $Q_{80}$, $Q_{90}$ as QST bases, and separate training images into $\mathcal{D}_{base}$ and $\mathcal{D}_{meta}$. ResNet50 \cite{He2016Deepresnet} is adopted as the baseline model. The $\theta_{rem}$ is initialized with the corresponding parameters of model pretrained on ImageNet, and the $\theta_{bnb}$ are set as $\theta^{1}_{bnb} = \theta^{2}_{bnb} = ... = \theta^{M}_{bnb}$. In base training, we use SGD with Nesterov momentum update $\theta_{bnb}$ and $\theta_{rem}$. The learning rate is set to 0.0001, the minibatch to $128$, and the base training epochs to 10. In the meta-learner training, we use Adam to update $\theta_{meta}$, and the meta-optimization step size $\beta$ is fixed to $0.001$, the meta learning rate $\gamma$ to $0.004$, the meta-learner training epochs $ite_2$ to 20. All input images are pre-processed with standard random cropping horizontal mirroring.
\subsubsection{Results on ImageNet and ImageNet-J}
We evaluate QAM on different domains of ImageNet. Succinctly, testing images are separated into 5 domains, including $Q_{96}, Q_{75}, Q_{99}, Q_{90}, Q_{others}$. As shown in Table \ref{tab:results imagenet}, QAM achieves $0.8\%$ improvement than the baseline model. Moreover, to measure the resilience of QAM to images with various levels of compression loss, we evaluate QAM on ImageNet-J. As Table \ref{tab:imagenet-j} shows, QAM performs better on images containing heavier compression loss, which differs from CIFAR-10-J and CIFAR-100-J. The reason is that most images in ImageNet are compressed slightly, as shown in Fig. \ref{fig:im_dist}. As a result, the baseline performs poorly on images containing heavy compression loss. Meanwhile, we make a Gradient-weighted Class Activation Mapping \cite{selvaraju2017grad} of the baseline and QAM. As shown in Fig. \ref{fig:gradmap}, suffering from even slight compression loss, the baseline may not focus on target objects but the background, while QAM could focus on the parts of target objects.

\subsubsection{Generalization Capability Evidence}
To investigate the generalization capability of QAM to other networks, we apply QAM to two state-of-the-art classification networks that utilize batch normalization, including a RegNet-16GF \cite{regnetradosavovic2020designing} and an EfficientNetV2-S \cite{tan2021efficientnetv2}.  We could observe from Table \ref{tab:regnet} that QAM achieves significant improvements on these two networks, which evidences the effectiveness of QAM.

\subsubsection{Comparison with Domain Generalization}
We compare QAM with the recent state-of-the-art domain generalization methods, and the results are shown in Table \ref{tab:imagenet-dg}. We observe that these domain generalization methods perform more poorly than the baseline shown in Table \ref{tab:results imagenet}. The reason for this phenomenon is that domains in ImageNet are \textit{imbalanced} (as shown in Fig. \ref{fig:im_dist}, the $Q_{96}$ domain accounts for nearly $80\%$). These domain generalization methods do not consider the problem of large differences in the amount of data and equally sample images from each domain. The results confirm the effectiveness of QAM on imbalanced data and also indicate that the problem discussed in this manuscript cannot be simply treated as a domain generalization problem.
\begin{table}[!tb]
\centering
\caption{The performance of ResNet-50 for different QSTs on ImageNet.}
\small
\label{tab:results imagenet}
{
\begin{adjustbox}{max width=\linewidth}
\begin{tabular}{ccccccc}
\toprule
ImageNet &$Q_{96}$& $Q_{75}$& $Q_{80}$& $Q_{90}$& $Q_{Others}$&  Avg. \\
\midrule
Baseline    &75.9          & 79.1          & 76.9          & 77.6 & 75.9          & 76.0   \\
QAM (ours)    &\textbf{76.7} & \textbf{79.7} & \textbf{77.0} & \textbf{77.7} & \textbf{76.8} & \textbf{76.8}\\
\bottomrule
\end{tabular}
\end{adjustbox}
}
\vspace{-3mm}
\end{table}
\begin{table}[!tb]
\caption{Comparison with state-of-the-art domain generalization methods on ImageNet.}
\centering
\small
\label{tab:imagenet-dg}
\begin{adjustbox}{max width=\linewidth}
\begin{tabular}{ccccccc}
\toprule
ImageNet &$Q_{96}$& $Q_{75}$& $Q_{80}$& $Q_{90}$& $Q_{Others}$&  Avg. \\
\midrule
SD \cite{SDpezeshki2020gradient}         & 69.9          & 76.0          & 69.1          & 71.8          & 72.8          & 70.4          \\
SagNet \cite{SagNetnam2021reducing}       & 68.0          & 76.7          & 69.0          & 71.2          & 71.4          & 68.7          \\
SelfReg \cite{kim2021selfreg}      & 68.6          & 75.4          & 68.8          & 73.1          & 72.4          & 69.3          \\
MLDG \cite{mldgli2018learning}         & 68.9          & 75.8          & 71.0          & 72.6          & 72.0          & 69.5          \\
MMD \cite{mmdli2018domain}          & 69.3          & 73.0          & 70.5          & 71.2          & 70.8          & 69.6          \\
CORAL \cite{coralsun2016deep}        & 68.9          & 77.0          & 70.5          & 72.5          & 72.3          & 69.6          \\
MTL \cite{mtlblanchard2017domain}          & 68.6          & 74.8          & 71.0          & 70.7          & 71.3          & 69.2          \\
Fish \cite{fishshi2021gradient}         & 69.0          & 75.4          & 69.6          & 72.5          & 71.6          & 69.5          \\
QAM (ours) & \textbf{76.7} & \textbf{79.4} & \textbf{77.1} & \textbf{77.7} & \textbf{76.9} & \textbf{76.8}\\
\bottomrule
\end{tabular}
\end{adjustbox}
\vspace{-3mm}
\end{table}
\begin{table}[tb!]
\caption{Accuracy of QAM on images with QSTs that are not produced by the default quantization tables.}
\centering
\label{tab:nonstandard}
\begin{adjustbox}{max width=\linewidth}
\begin{tabular}{ccccc}
\toprule
QST & $Q_{u1}$\cite{chao2013design}          & $Q_{u2}$\cite{tuba2014jpeg}  & $Q_{u3}$\cite{tuba2017jpeg}     & $Q_{u4}$\cite{duan2012optimizing}       \\
\midrule
Baseline & 72.6   &69.3     & 72.6  & 65.9       \\
QAM & \textbf{73.6} &\textbf{70.4} & \textbf{74.3} & \textbf{67.2} \\
\bottomrule
\end{tabular}
\end{adjustbox}
\vspace{-3mm}
\end{table}

\subsubsection{Results on QSTs that are not produced by the default quantization tables}
As mentioned in Section \ref{sec:jpeg}, JPEG allows the quantization tables to be customized at the encoder. In this section, we evaluate QAM on QSTs that are not produced by the default quantization tables. Specifically, we compress each test image in ImageNet using 4 quantization tables, including the quantization tables designed in \cite{chao2013design}, \cite{tuba2014jpeg}, \cite{tuba2017jpeg}, and \cite{duan2012optimizing}. We denote the corresponding QSTs as $Q_{u1}$, $Q_{u2}$, $Q_{u3}$ and $Q_{u4}$ respectively. The results are shown in Table \ref{tab:nonstandard}. It could be observed that QAM achieves $1.0\%$, $1.1\%$, $1.7\%$, and $1.3\%$ improvements over the baseline, respectively. The results demonstrate that QAM is also effective on images that are not compressed with the default quantization tables.

\section{Conclusion}
\label{sec:con}
The sensitivity of deep neural networks to compressed images causes unreasonable phenomena and prevents the networks from being deployed in safety-critical applications. In this paper, we propose a novel perspective, utilizing one of the disposable coding parameters, quantization steps, to reduce the sensitivity of the networks to compressed images and improve the performance of image classification. Experimental results demonstrate the effectiveness and generalization of the proposed method. Moreover, it is worth extending our work to video coding, where more complex compression algorithms are applied, and more coding parameters could be picked up. On the other hand, some classification networks like Vision Transformers (ViT) do not employ batch normalization. Thus, the proposed method could not be used in these networks, and how to serve the networks without batch normalization is one of our future directions.

\ifCLASSOPTIONcompsoc
  \section*{Acknowledgments}
\else
  \section*{Acknowledgment}
\fi
This work was supported by Key-Area Research and Development Program of Guangdong Province under Contract 2033811500015, and in part by the National Natural Science Foundation of China under Contract 62027804, Contract 61825101 and Contract 62088102. The computing resources of Pengcheng Cloudbrain are used in this research. 

\ifCLASSOPTIONcaptionsoff
  \newpage
\fi



\bibliographystyle{IEEEtran}
\bibliography{ma_tcsvt.bib}
%

%
\begin{IEEEbiography}[{\includegraphics[width=1.0in,height=1.25in,clip,keepaspectratio]{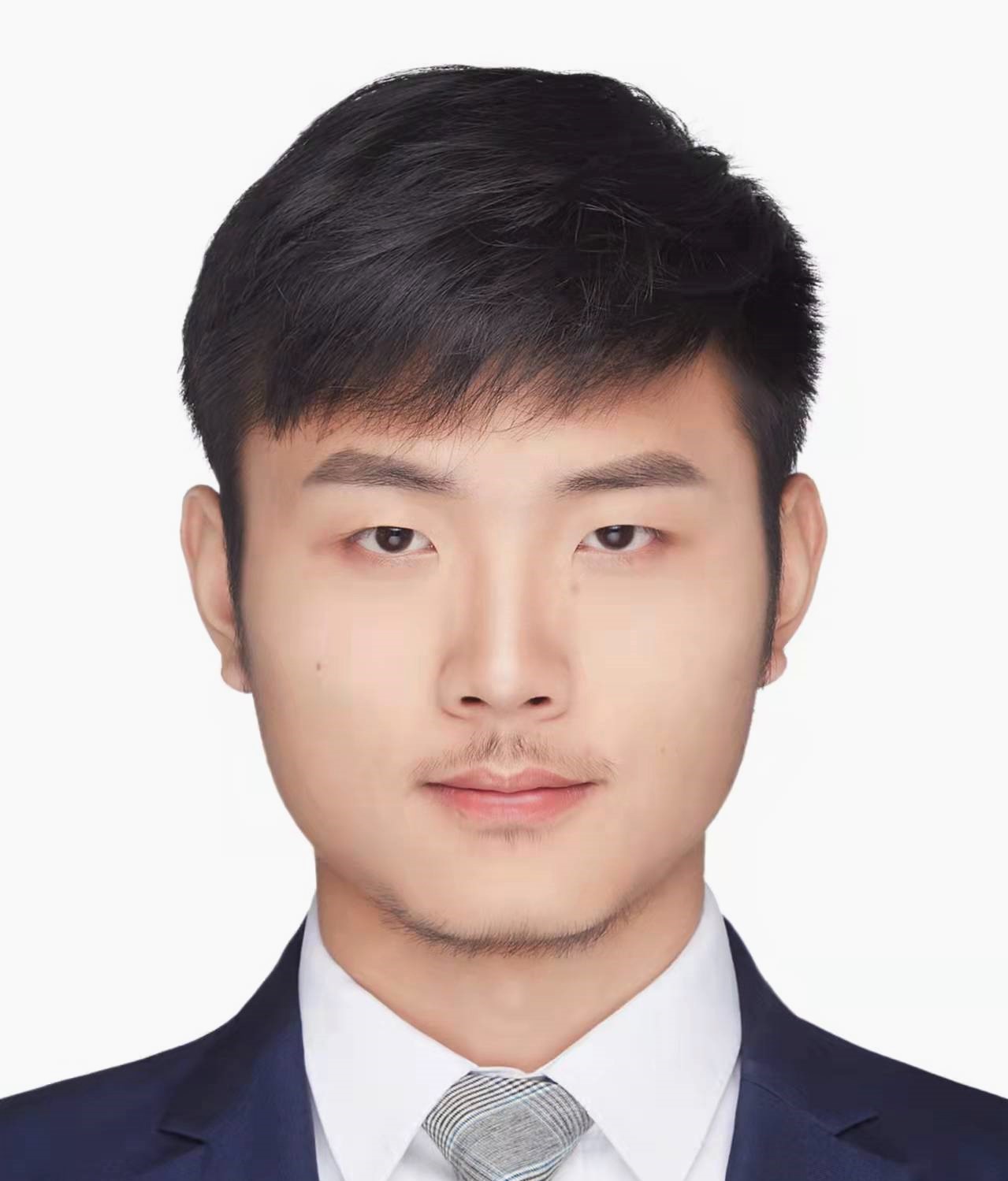}}]{Li Ma} received the B.S. degree from Peking University, Beijing, China, in 2016. He is pursuing a Ph.D. candidate in School of Computer Science, Peking University, China. His research interests include image compression, video coding and multimedia big data.
\end{IEEEbiography}
\vspace{-15mm}
\begin{IEEEbiography}[{\includegraphics[width=1.0in,height=1.25in,clip,keepaspectratio]{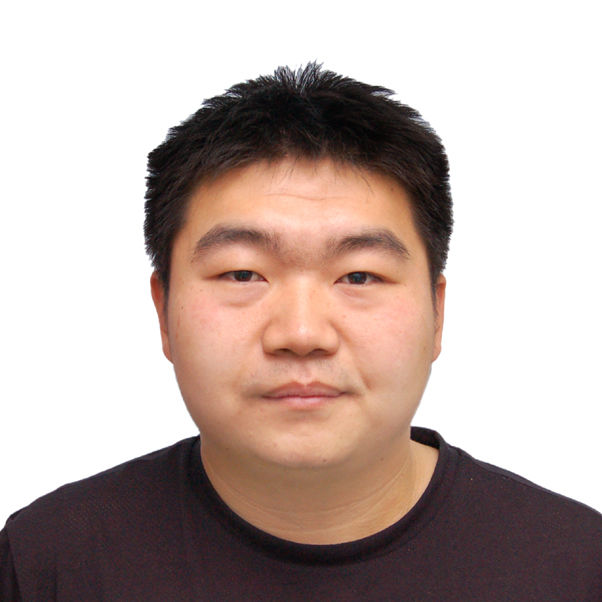}}]{Peixi Peng} received the PhD degree from  Peking University, in 2017, Beijing, China. He is currently a associate researcher with the School of Computer Science, Peking University, Beijing, China, and is also the assistant researcher of Artificial Intelligence Research Center, Peng Cheng Laboratory, Shenzhen, China.  He is the author or co-author of more than 20 technical articles in refereed journals such as IEEE TPAMI, PR and conferences such as CVPR/ECCV/IJCAI/ACMMM/AAAI. His research interests include computer vision, multimedia big data, and reinforcement learning.
\end{IEEEbiography}
\vspace{-13 mm}
\begin{IEEEbiography}[{\includegraphics[width=1.0in,height=1.25in,clip,keepaspectratio]{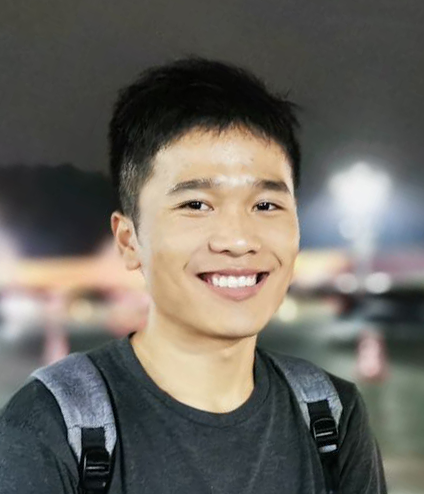}}]{Guangyao Chen} received a BS degree from Wuhan University, in 2018, Wuhan, China.
He is pursuing a PhD degree in School of Computer Science, Peking University, China. His research interests include open world vision, out-of-distribution, and neural network compression.
\end{IEEEbiography}
\vspace{-15 mm}
\begin{IEEEbiography}[{\includegraphics[width=1.0in,height=1.25in,clip,keepaspectratio]{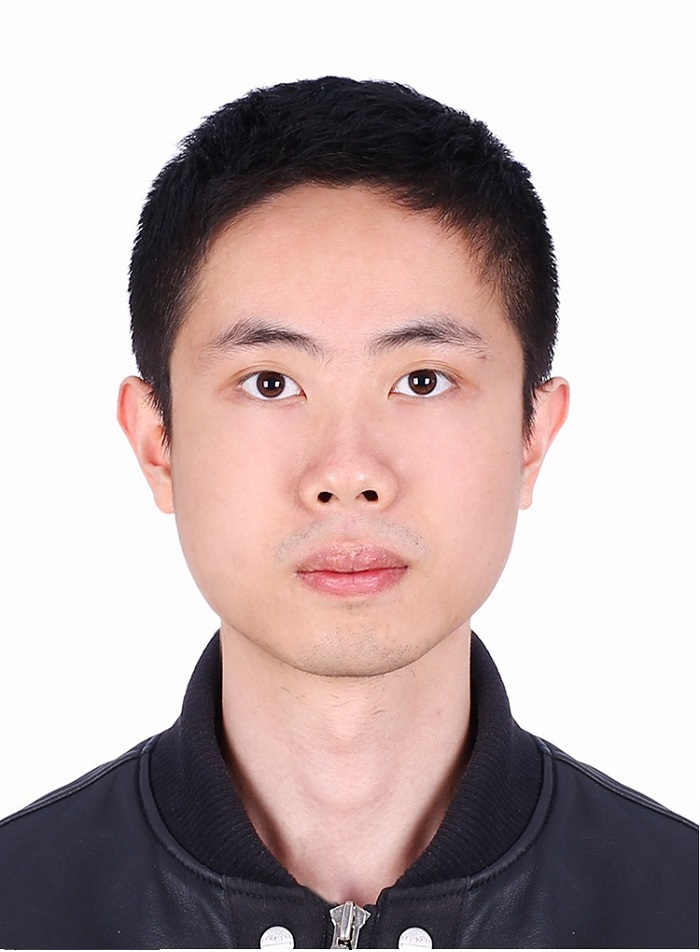}}]{Yifan Zhao} is currently a Postdoc researcher with School of Computer Science, Peking University, Beijing, China. He received the B.E. degree from Harbin Institute of Technology in Jul. 2016 and the Ph.D. degree from School of Computer Science and Engineering, Beihang University, in Oct. 2021. His research interests include computer vision and image/video understanding.
\end{IEEEbiography}
\vspace{-13 mm}
\begin{IEEEbiography}[{\includegraphics[width=1.0in,height=1.25in,clip,keepaspectratio]{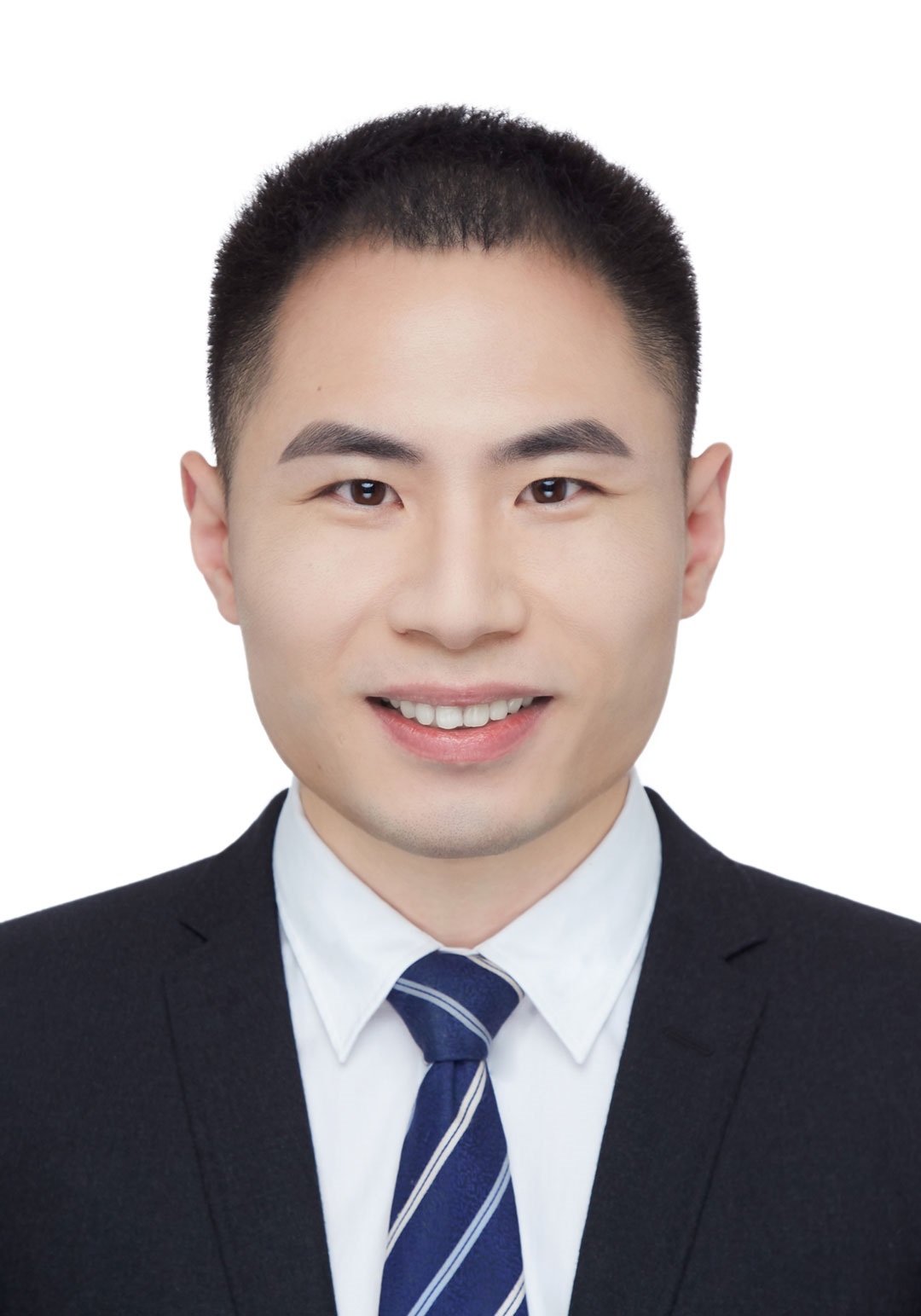}}]{Siwei Dong} received the Ph.D. degree at the School of Electronics Engineering and Computer Science, Peking University, Beijing, China, in 2019. Previously, He received the B.S. degree from Chongqing University, Chongqing, China, in 2012. He is currently a researcher at Beijing Academy of Artificial Intelligence (BAAI). His research interests include healthcare computing, video coding and neuromorphic computing.
\end{IEEEbiography}
\begin{IEEEbiography}[{\includegraphics[width=1.0in,height=1.25in,clip,keepaspectratio]{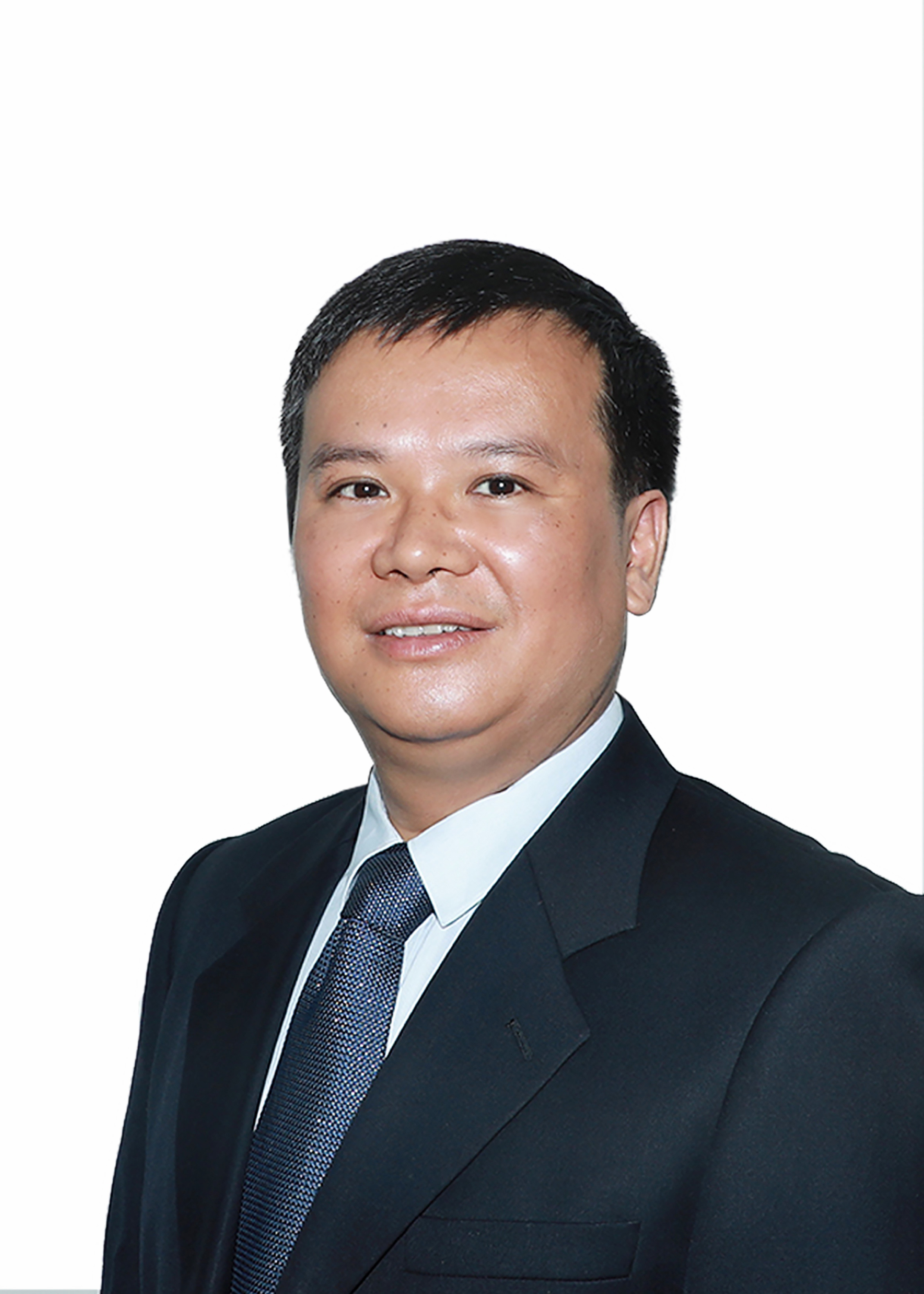}}]{Yonghong Tian}(S’00-M’06-SM’10-F’22) is currently the Dean of School of Electronics and Computer Engineering, a Boya Distinguished Professor with the School of Computer Science, Peking University, China, and is also the deputy director of Artificial Intelligence Research Department, PengCheng Laboratory, Shenzhen, China. His research interests include neuromorphic vision, distributed machine learning and multimedia big data. He is the author or coauthor of over 300 technical articles in refereed journals and conferences. Prof. Tian was/is an Associate Editor of IEEE TCSVT (2018.1-2021.12), IEEE TMM (2014.8-2018.8), IEEE Multimedia Mag. (2018.1-2022.8), and IEEE Access (2017.1-2021.12). He co-initiated IEEE Int’l Conf. on Multimedia Big Data (BigMM) and served as the TPC Co-chair of BigMM 2015, and aslo served as the Technical Program Co-chair of IEEE ICME 2015, IEEE ISM 2015 and IEEE MIPR 2018/2019, and General Co-chair of IEEE MIPR 2020 and ICME2021. He is the steering member of IEEE ICME (2018-2020) and IEEE BigMM (2015-), and is a TPC Member of more than ten conferences such as CVPR, ICCV, ACM KDD, AAAI, ACM MM and ECCV. He was the recipient of the Chinese National Science Foundation for Distinguished Young Scholars in 2018, two National Science and Technology Awards and three ministerial-level awards in China, and obtained the 2015 EURASIP Best Paper Award for Journal on Image and Video Processing, and the best paper award of IEEE BigMM 2018, and the 2022 IEEE SA Standards Medallion and SA Emerging Technology Award. He is a Fellow of IEEE, a senior member of CIE and CCF, a member of ACM.
\end{IEEEbiography}

\end{document}